\documentclass[10pt,twocolumn,letterpaper]{article}

\usepackage{cvpr}              

\usepackage[dvipsnames]{xcolor}
\usepackage{graphicx}
\usepackage{amsmath}
\usepackage{amssymb}
\usepackage{pifont}
\usepackage{booktabs}
\usepackage{colortbl}
\usepackage{enumitem}
\usepackage{subcaption}

\usepackage{svg}
\usepackage{pgf-pie}

\graphicspath{{figs/}}

\definecolor{cvprblue}{rgb}{0.21,0.49,0.74}
\usepackage[pagebackref,breaklinks,colorlinks,citecolor=cvprblue]{hyperref}

\usepackage{nicematrix}
\usepackage[capitalize]{cleveref}
\crefname{section}{Sec.}{Secs.}
\Crefname{section}{Section}{Sections}
\Crefname{table}{Table}{Tables}
\crefname{table}{Tab.}{Tabs.}
\Crefformat{section}{\S#2#1#3}
\Crefformat{figure}{#2Fig.~#1#3}
\Crefformat{table}{#2Tab.~#1#3}

\newcommand{\xmark}{\textcolor[HTML]{e74c3c}{\ding{55}}}
\newcommand{\cmark}{\textcolor{teal}{\bf \ding{51}}}

\newcommand{\method}{BEHAVIOR Vision Suite\xspace}
\newcommand{\methodabbr}{BVS\xspace}
\newcommand{\asset}{extended BEHAVIOR-1K assets\xspace}
\newcommand{\generator}{customizable dataset generator\xspace}

\definecolor{cvprblue}{rgb}{0.21,0.49,0.74}

\def\paperID{10394} 
\def\confName{CVPR}
\def\confYear{2024}

\author{Yunhao Ge$^{1,2*}$, Yihe Tang$^{1*}$, Jiashu Xu$^{3*}$, Cem Gokmen$^{1*}$, Chengshu Li$^{1}$, Wensi Ai$^{1}$, \\
Benjamin Jose Martinez$^{1}$, Arman Aydin$^{1}$, Mona Anvari$^{1}$, Ayush K Chakravarthy$^{1}$, Hong-Xing Yu$^{1}$, \\
Josiah Wong$^{1}$, Sanjana Srivastava$^{1}$, Sharon Lee$^{1}$, Shengxin Zha$^{4}$, Laurent Itti$^{2}$, Yunzhu Li$^{1,7}$, \\
Roberto Martín-Martín$^{6}$, Miao Liu$^{4}$, Pengchuan Zhang$^{5}$, Ruohan Zhang$^{1}$, Li Fei-Fei$^{1}$, Jiajun Wu$^{1}$\\
\\
$^{1}$Stanford University \ \  $^{2}$University of Southern California \ \ $^{3}$Harvard University\ \  $^{4}$GenAI, Meta \\ $^{5}$FAIR, Meta \ \
$^{6}$The University of Texas at Austin \ \ 
$^{7}$The University of Illinois Urbana-Champaign
}

\begin{document}
\title{\method: Customizable Dataset Generation via Simulation}



\maketitle

\begin{abstract}
\let\thefootnote\relax\footnotetext{$*$ equal contribution}
\let\thefootnote\relax\footnotetext{$^\dagger$ correspondence to \texttt{yunhaoge@cs.stanford.edu},\\ \texttt{\{yihetang, zharu\}@stanford.edu}}
The systematic evaluation and understanding of computer vision models under varying conditions require large amounts of data with comprehensive and customized labels, which real-world vision datasets rarely satisfy. While current synthetic data generators offer a promising alternative, particularly for embodied AI tasks, they often fall short for computer vision tasks due to low asset and rendering quality, limited diversity, and unrealistic physical properties. 
We introduce the \method (\methodabbr), a set of tools and assets to generate fully customized synthetic data for systematic evaluation of computer vision models, based on the newly developed embodied AI benchmark, BEHAVIOR-1K.
\methodabbr supports a large number of adjustable parameters at the scene level (e.g., lighting, object placement), the object level (e.g., joint configuration, attributes such as ``filled'' and ``folded''), and the camera level (e.g., field of view, focal length). Researchers can arbitrarily vary these parameters during data generation to perform controlled experiments. 
We showcase three example application scenarios: systematically evaluating the robustness of models across different continuous axes of domain shift, evaluating scene understanding models on the same set of images, and training and evaluating simulation-to-real transfer for a novel vision task: unary and binary state prediction. Project website: \textcolor{blue}{\url{https://behavior-vision-suite.github.io/}}

\end{abstract}

\section{Introduction}
\label{sec:intro}

\begin{figure*}
\begin{center}
\includegraphics[width=0.8\linewidth]{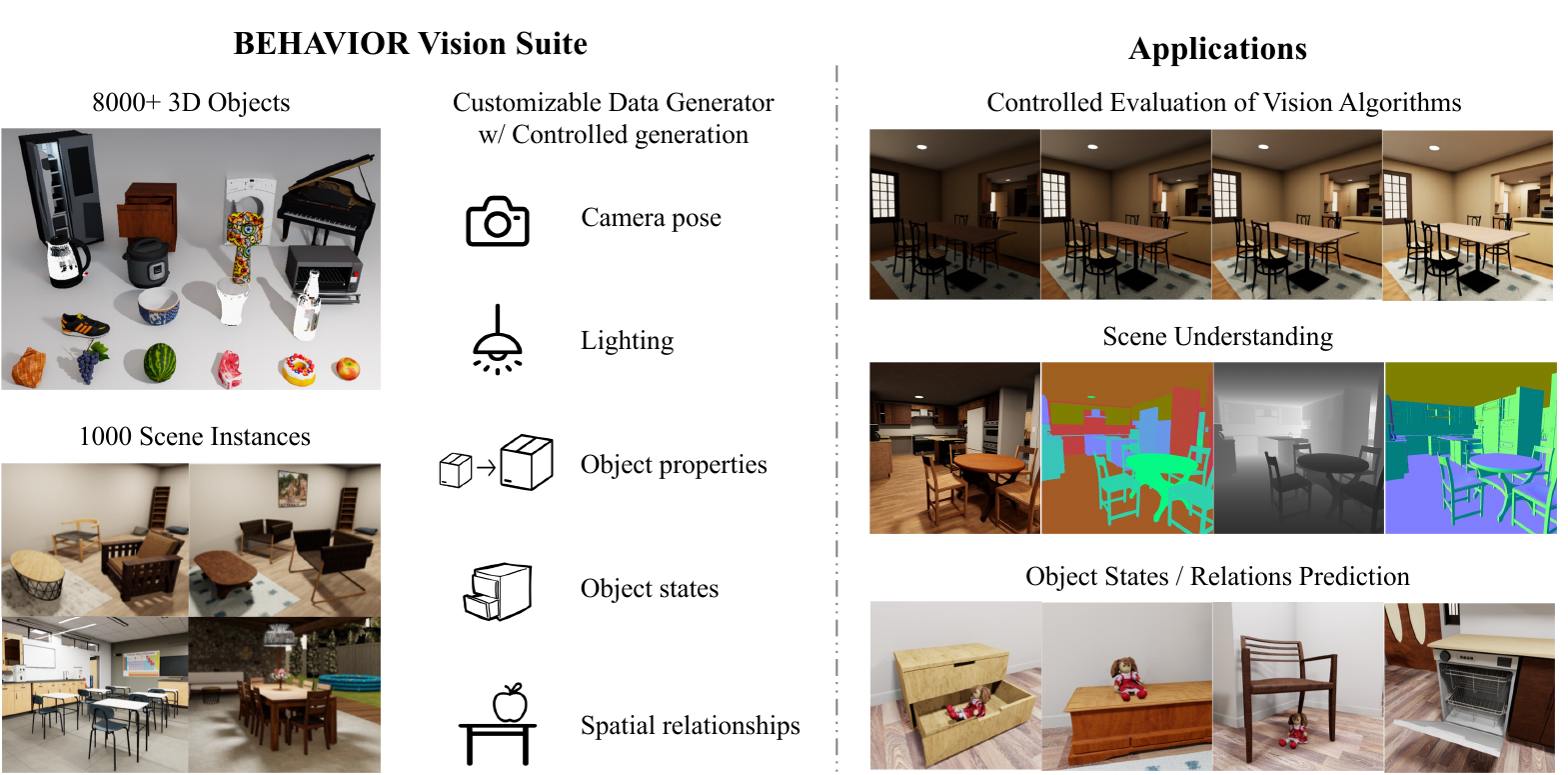}
\end{center}
\vspace{-15pt}
    \caption{Overview of \method (\methodabbr), our proposed toolkit for computer vision research. \methodabbr builds upon the extended object assets and scene instances from BEHAVIOR-1K \cite{li2023behavior1k}, and provides a customizable data generator that allows users to generate photorealistic, physically plausible labeled data in a controlled manner. We demonstrate \methodabbr with three representative applications.}
\vspace{-15pt}
\label{fig:capgen}
\end{figure*}



Large-scale datasets and benchmarks have fueled computer vision research in the past decade~\cite{deng2009imagenet,lin2014microsoft,everingham2010pascal,krishna2017visual,geiger2012we,goyal2017something,sigurdsson2018actor,xiang2017posecnn,martin2021jrdb,caba2015activitynet,gurari2018vizwiz, grauman2022ego4d}. Driven by these datasets and benchmarks, thousands of models and algorithms tackling different perception challenges are being proposed every year, on the topics of object detection~\cite{zou2023object}, segmentation~\cite{kirillov2023segment}, action recognition~\cite{wang2013action}, video understanding~\cite{lin2019tsm} and beyond.
Despite their success, real-world datasets face inherent limitations. First, the ground-truth object/pixel-level labels are either prohibitively expensive to acquire (e.g., segmentation masks)~\cite{cocodataset} or suffering from inaccuracies (e.g., depth sensing)~\cite{NYUv2}. Consequently, each real dataset often only offers limited labels,  hindering the development and evaluation of computer vision models that perform a wide range of perception tasks on the same input. Even when annotations are affordable and accurate, real-world datasets are limited by the availability of source images. For example, images of rare events, such as traffic accidents or low-light conditions, might be difficult to acquire from the Internet or real-world sensors. Finally, once collected, these real-world datasets have a fixed data distribution and cannot be easily changed. This makes it challenging for researchers to conduct customized experiments, 
often leading to models that overfit the datasets and eventually rendering the entire benchmarks obsolete ~\cite{hendrycks2021natural, hendrycks2021many,shen2021igibson, li2023behavior1k}.

To avoid this limitation, researchers and practitioners have devised various methods to generate synthetic datasets that complement the real ones \cite{ge2023beyond}. In the realm of indoor scene understanding, 3D reconstruction datasets~\cite{xiazamirhe2018gibsonenv, Matterport3D, ramakrishnan2021habitat} provide a promising avenue to generate source images from arbitrary viewpoints and free (geometric) annotations. However, due to the imperfect nature of 3D reconstruction techniques, the rendered images are not very realistic. Since each entire scene is a static mesh, these datasets offer very limited customizability beyond camera trajectories. Recent synthetic indoor datasets (often designed by 3D artists)~\cite{roberts2021hypersim, li2018interiornet, li2020openrooms, fu20213d} not only offer free geometric and semantic annotations, but also support object layout reconfiguration as objects are usually independent CAD models. However, these datasets do not guarantee physical plausibility, as object penetration and levitation occur frequently, and offer no customization capability beyond changing object poses. 3D simulators~\cite{szot2021habitat, kolve2017ai2, deitke2022️, gan2020threedworld, shen2021igibson, li2021igibson}, on the other hand, guarantee physical plausibility with their underlying physics engines. They allow users to customize the joint configuration of articulated objects and even more advanced object states such as ``cooked'' or ``sliced''~\cite{kolve2017ai2,li2021igibson}. Yet these 3D simulators generally cater to embodied AI and robotics researchers, and as a result, they lack photorealism compared to the synthetic datasets mentioned before (often due to speed constraints), and do not offer ready-made tools to generate customized image/video datasets for computer vision researchers. 

To overcome the aforementioned challenges, we propose \method (\methodabbr), a customizable data generation tool that enables systematic evaluation and understanding of computer vision models (see \Cref{fig:capgen} for an overview). To do so, we expand the 3D asset library in BEHAVIOR-1K~\citep{li2023behavior1k}, focusing on enhancing both object diversity and scene variety, as well as adding features to increase the value of the assets for vision tasks. We also introduce a \generator, which uses the simulator from the BEHAVIOR-1K benchmark~\cite{srivastava2022behavior, li2023behavior1k} to generate custom vision datasets. We build a versatile and customizable toolbox to generate high-quality synthetic data for systematic model evaluation and understanding. 

In summary, \method possesses the following unique combination of desirable features:
\begin{itemize}
    \item \methodabbr offers image/object/pixel-level labels (scene graph, point cloud, depth, segmentation, etc.);
    
    \vspace{0.05cm}
    \item \methodabbr covers a wide variety of indoor scenes and objects (8K+ objects, 1K scene instances, fluid, soft bodies);
    
    \vspace{0.05cm}
    \item \methodabbr provides physical plausibility and photorealism;
    
    \vspace{0.05cm}
    \item \methodabbr supports customization in terms of object models, poses, joint configurations, semantic states, lighting, texture, material, camera setting, etc.;
    
    \vspace{0.05cm}
    \item \methodabbr includes easy-to-use tooling to generate customized data for new use cases.
\end{itemize}

To demonstrate the usefulness of \methodabbr, we show three example applications: 1) parametrically evaluating model robustness across different conditions such as lighting and occlusion, 2) evaluating different types of representative computer vision models on the same set of images, and 3) training and evaluating sim2real transfer for object state and relation prediction. We hope that \methodabbr can unlock more possibilities for the computer vision community.

\section{Related works}
\label{sec:relatedwork}

In this section, we compare \method with other real RGB-D datasets, 3D reconstruction datasets, synthetic datasets, and 3D simulators in terms of customizability and visual quality (see \Cref{tab:comparison_main}). 
\begin{table}
  \scriptsize \centering 
  \setlength{\tabcolsep}{2.5pt}
  \begin{NiceTabular}[baseline=1,cell-space-limits=0pt]{lccccc} 
  \CodeBefore
     \rectanglecolor{gray!30}{4-1}{6-6}
  \Body
  \toprule 
    \Block{2-1}{Dataset Category} & \Block{1-4}{Customizability}  & &  & &\Block{2-1}{Visual \\ Quality} \\
    \cmidrule{2-5}
     & Camera & Obj.~Pose & Obj.~State & Toolkit \\ 
     \midrule
    Real RGB-D Datasets & \xmark & \xmark & \xmark & N/A & Real \\ 
    \midrule
    3D Reconstruction Datasets & \cmark & \xmark & \xmark & N/A & Medium \\ 
    Synthetic Datasets & \cmark & \cmark & \xmark & {\color{orange}$\sim$} & High \\
    3D Simulators & \cmark & \cmark & \cmark & \xmark & Low \\
    \midrule
    \method (ours) & \cmark & \cmark & \cmark & \cmark & High \\
    \bottomrule
  \end{NiceTabular}
  \vspace{-6pt}
  \caption{Comparison of real and different types of \colorbox{gray!30}{synthetic} datasets with \method. `Camera' denotes the ability to render images from any viewing angle. `Obj.~Pose' refers to the modifiability of the object layout. `Obj.~State' indicates whether an object's physical states (e.g., open/close, folded) and semantic states (e.g., cooked, soaked) can be modified. `Toolkit' indicates the availability of utility functions for sampling object layout and camera poses under specified conditions (e.g., viewing half-open kitchen cabinets filled with grocery items). `Visual Quality' evaluates the photorealism of rendered images. 
  }
  
  \vspace{-10pt}
  \label{tab:comparison_main}
\end{table}

\vspace{2pt}
\noindent\textbf{Real Indoor Scene RGB-D Datasets.}
RGB-D image datasets of real indoor scenes~\cite{NYUv2, Song_2015_CVPR, dai2017scannet, dehghan2021arkitscenes, yeshwanthliu2023scannetpp} have driven advances in 3D perception and holistic scene understanding, with recent additions like ARKitScenes~\cite{dehghan2021arkitscenes} and ScanNet++~\cite{yeshwanthliu2023scannetpp} offering dense semantic and 3D annotations. Despite having minimum domain gaps with respect to real-world applications, these real datasets are expensive to annotate and inherently static, limiting users' ability to generate images from novel camera views, acquire new types of annotations, or alter scenes. Our work complements these limitations by offering a fully customizable generator for photorealistic synthetic data.

\vspace{2pt}
\noindent\textbf{3D Reconstruction Datasets.}
3D reconstruction datasets such as Gibson, Matterport, and HM3DSem~\cite{Matterport3D, xiazamirhe2018gibsonenv, ramakrishnan2021habitat, yadav2022habitat} allow the rendering of novel views.
While these datasets have tremendously benefited the embodied navigation community, their utility for broader computer vision applications remains limited. Each scene, being a single 3D mesh, restricts further customization, such as modifying the object layout. Moreover, the visual quality of rendered novel views depends on the reconstruction's fidelity, often resulting in artifacts. While Taskonomy~\cite{taskonomy2018} and Omnidata~\cite{eftekhar2021omnidata} have extended mid-level visual cues such as surface normal for these datasets, semantic label acquisition remains expensive. In contrast, our work offers the flexibility to generate images with customized object layouts with consistent visual quality, while also providing comprehensive labels at no additional cost.

\vspace{2pt}
\noindent\textbf{Synthetic Datasets.}
Synthetic datasets offer an alternative approach that eliminates the need for manual semantic labeling by rendering realistic images from interior scenes composed of independent object models~\cite{objaverse, objaverseXL}. Object layouts are usually created by artists~\cite{roberts2021hypersim, fu20213d, li2018interiornet} or parsed from real scans~\cite{li2020openrooms}, offering semantic realism of the scenes~\cite{wang2022neural, ge20243d}. Methods like OpenRooms~\cite{li2020openrooms} and Unity Synthetic Homes~\cite{SyntHomes} also allow users to configure rendering options, such as lighting. 
However, despite their photorealism, the rendered images often lack physical plausibility, with common issues like object penetration or slight levitation. In addition, the object models are mostly fully rigid and support very limited semantic states. 
Our generator not only ensures the physical plausibility of images created, but also supports broader relationship customization (e.g., ``cooked'' or ``filled'') and more granular control over the sampled state, such as openness level through joint limit annotations. 


\vspace{2pt}
\noindent\textbf{3D Simulators.}
A large number of 3D simulators with physical realism have been developed recently. 
Kubric~\cite{greff2021kubric} focuses on generating physically plausible object clusters without full-scene simulation. iGibson~\cite{li2021igibson, shen2021igibson} and Habitat 2.0~\cite{szot2021habitat} offer reconfigurable indoor scenes with articulated assets, the former notably supporting extended object states such as wetness level. ThreeDWorld~\cite{gan2020threedworld} emphasizes physical prediction, especially with non-rigid objects. ProcTHOR~\cite{deitke2022️} automates the large-scale generation of semantically plausible virtual environments. Since these 3D simulators cater to the embodied AI and robotics community, their visual quality is often not prioritized. In contrast, we use OmniGibson, a new simulator that surpasses the photorealism of the aforementioned ones, according to a user study~\cite{li2023behavior1k}, positioning our work as more suitable for computer vision research. 
Moreover, we provide a range of utility functions in this work, allowing for easy creation of diverse images tailored to specific needs---a feature most existing 3D simulators lack. 

\section{\method}
\label{sec:BVisionSuite}

\begin{figure*}[t]
\begin{center}
\includegraphics[width=\textwidth]{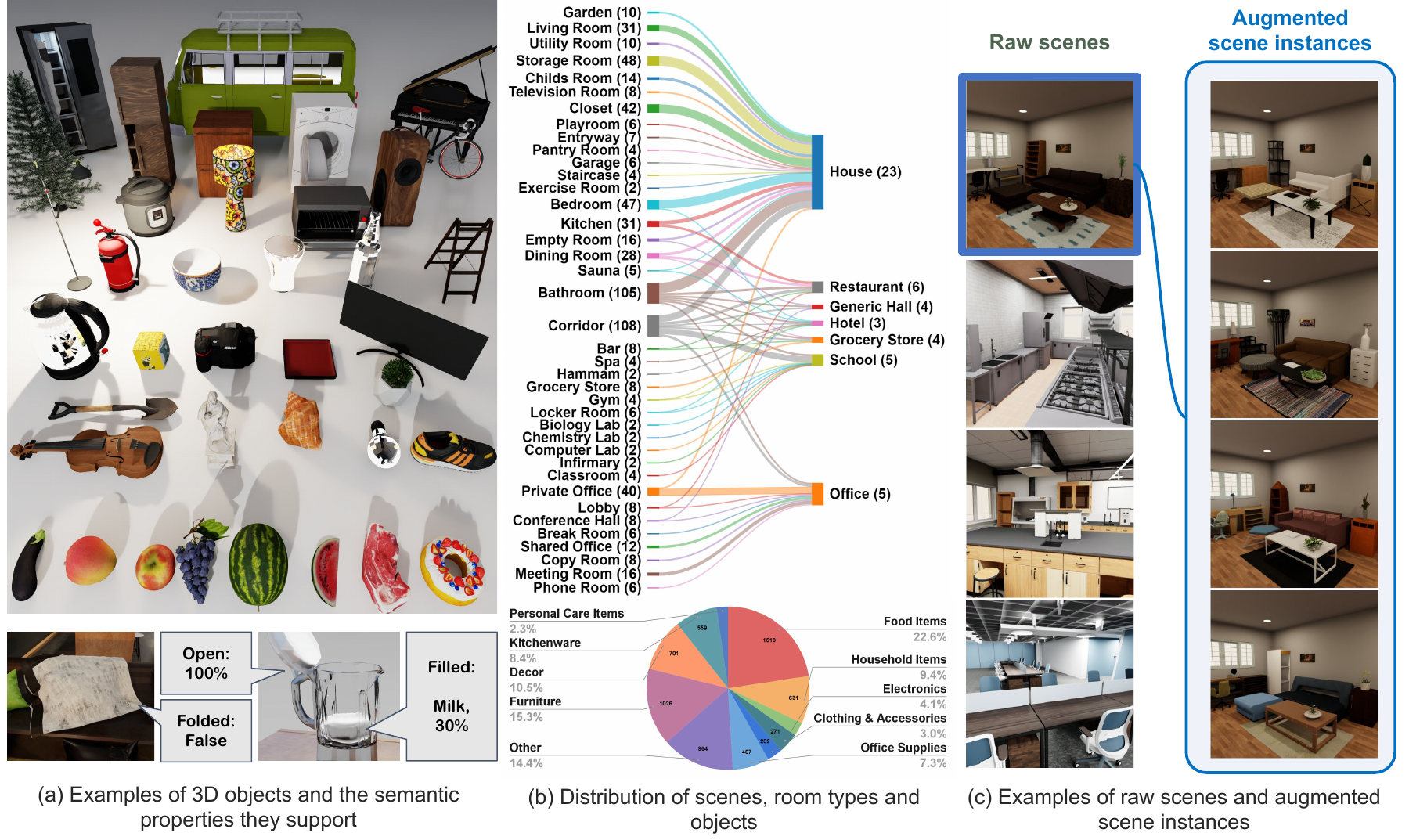}
\end{center}
      \vspace{-18pt} 
    \caption{Overview of \asset: Covering a wide range of object categories and scene types, our 3D assets have high visual and physical fidelity and rich annotations of semantic properties, allowing us to generate 1,000+ realistic scene configurations.}
    \vspace{-12pt}
\label{fig:asset}
\end{figure*}


\method contains two main components (\Cref{fig:capgen}): the \asset and the \generator. The assets serve as the foundation, while the generator leverages these assets to create vision datasets tailored to downstream tasks of interest.

\subsection{Extended BEHAVIOR-1K Assets}
\label{sec:BAsset}

The \asset comprises a diverse collection of 8,841 object models and 1,000 scene instances, derived from 51 artist-designed raw scenes. 
Of these objects, 2,156 are structural elements like walls, floors, and ceilings, while the remaining 6,685 nonstructural items span 1,937 categories, including food, tools, electronics, clothing, and office supplies, among others. This categorization is detailed in \Cref{fig:asset}. Predominantly indoor, the 51 raw scenes also incorporate outdoor elements such as gardens and encompass a wide variety of environments: houses (23), offices (5), restaurants (6), grocery stores (4), hotels (3), schools (5), and generic halls (4), as well as a simulated twin of a mock apartment in our research lab. This collection of assets is the result of a year-long effort to extend the BEHAVIOR-1K~\cite{li2023behavior1k} assets to enhance their applicability in computer vision. 

We expanded the object collection from 5,215 to 8,841 by adding more everyday objects, segmenting building structures into individual objects for more precise 3D bounding box labels, and procedurally generating sliced food. In addition, we have developed functionality that enables the generation of diverse scene variations by altering furniture object models and incorporating additional everyday objects. We will release 1000 scene instances augmented from the 51 raw scenes. 

To improve physical realism, we refined collision meshes using V-HACD~\cite{mamou2016vhacd} and CoACD~\cite{wei2022coacd}, manually selecting the best parameters to ensure a balance between physical accuracy, affordance preservation, and simulation efficiency. For more than 2,000 objects, where this method was insufficient, we manually designed their collision meshes.

We enhanced lighting realism by annotating actual light source objects, such as lamps and ceiling lights, to mimic real-world illumination. For more detailed semantic properties, we annotated appropriate container fillable volumes (e.g., cups, pots) and fluid source/sink locations (e.g., faucets, drains, sprayers), enabling us to spawn fluids in the scene realistically. Scene objects were annotated if they cannot be freely moved, e.g., when they physically support other objects. Cluttered objects were distinctly annotated, allowing them to be replaced with alternative clutters.

Altogether, we designed the assets to form a strong basis for custom data generation (discussed in \Cref{sec:BEHAVIOR-Dataset-Generator}), with a functional organization that allows accurate object randomization, and the annotations to provide a large number of modifiable parameters at both the object and scene levels.

\subsection{Customizable Dataset Generator}
\label{sec:BEHAVIOR-Dataset-Generator}

The \generator, the software component of the \method, is designed to generate synthetic datasets tailored to particular specifications. Built on OmniGibson~\cite{li2023behavior1k}, it leverages NVIDIA Omniverse's photorealistic, real-time renderer and OmniGibson's procedural sampling functions for object states to generate custom images and videos that satisfy arbitrary requirements. 
The produced datasets include rich, comprehensive annotations--segmentation masks, 2D/3D bounding boxes, depth, surface normals, flows, and point clouds--at no additional cost. 
Crucially, it empowers users with extensive control over the dataset generation process, allowing them to specify requirements on scene layouts, object states, camera angles, and lighting conditions, all while ensuring physical plausibility through the physics engine.

\vspace{2pt}
\noindent\textbf{Capabilities.}
The generator has the following capabilities:
\begin{itemize}

\vspace{2pt}
\item \textit{Scene Object Randomization:} It can swap scene objects with alternative models from the same category, which are grouped based on visual and functional similarities. This randomization significantly varies scene appearances while maintaining layouts' semantic integrity. 

\vspace{2pt}
\item \textit{Physically Realistic Pose Generation:} The generator can procedurally change the physical states of objects to satisfy certain predicates. This includes 1) placing objects with respect to other objects in the scene in a certain way (e.g., inside, on top of, or under), 2) opening or closing articulated objects, 3) filling containers with fluids, and 4) folding or unfolding pieces of cloth. The generator can generate multiple valid configurations for the same predicate and ensures physical plausibility. 

\vspace{2pt}
\item \textit{Predicate-Based Rich Labeling:} Beyond usual labels (semantic \& instance segmentation, bounding boxes, surface normals, depth, etc.), the generator also provides annotations including unary states of an object (e.g., whether an articulated object is open, or an appliance is toggled on), binary predicates between two objects (e.g., if one is touching, on top of, next to another) or between an object and a substance (e.g., if an object is filled/covered/soaked with a substance), and continuous labels (e.g., joint openness for articulated objects, filled fraction for containers).

\vspace{2pt}
\item \textit{Camera Pose and Trajectory Sampling:} Finding proper camera pose in a 3D scene is a challenging but crucial step in the rendering pipeline: the camera shall not be occluded and points at the subject of interest. The generator uses occupancy grids and hand-crafted heuristics to generate both static camera poses and plausible traversal trajectories that satisfy these constraints to curate image or scene traversal video datasets.

\vspace{2pt}
\item \textit{Configurable Rendering:} Through a user-friendly API, the generator allows for the customization of rendering parameters, including lighting and camera specifics such as aperture and field of view. 
\end{itemize}

\vspace{2pt}
\noindent\textbf{Dataset Generation.}
Images in the \methodabbr dataset can be generated as follows.
  First, we select one of the 51 raw scenes from the user-configured scene category (say, an office). Scene objects are randomized with instances from the same category.
  Depending on the user configuration, we determine additional objects to add to the scene. We place the objects using the pose generation capabilities based on user-specified requirements. This might include cluttering certain areas (e.g., filling a fridge with perishables) or individually manipulating object states (e.g., making a cabinet open or a table covered with water) for predicate prediction.

  We then generate a camera pose (or a sequence of poses as a camera trajectory), as well as randomize the scene's lighting parameters and the camera's intrinsics based on the user's specifications.
  Finally, we render an image (or a sequence of images) and record it alongside all relevant labels requested by the user, including additional modalities (depth/segmentation/etc.), bounding boxes, and predicate and object state values.

\section{Applications and Experiments}
\label{sec:Use CDG}

We present three applications and corresponding experiments to demonstrate the utility of \methodabbr: first, systematically evaluating model robustness against various continuous domain shifts, such as the lighting condition (\Cref{sec:App-PME}); second, assessing various scene understanding models using a consistent set of images with comprehensive annotations (\Cref{sec:App-HSU}); and third, training a model for a new vision task, object states and relations prediction, on synthesized data and evaluating its simulation-to-real transfer capability. (\Cref{sec:App-UB}). 
\subsection{Parametric Model Evaluation}
\label{sec:App-PME}


\begin{table}[t]
    \small
    \centering
    \resizebox{0.65\linewidth}{!}{
    \begin{NiceTabular}[baseline=2,cell-space-limits=1pt]{lcc} \toprule
        Axis & \# scenes & \# video clips \\
        \midrule
        Articulation & 17 & 237 \\
        Lighting & 16 & 441\\
        Visibility & 14 & 211\\
        Zoom & 9 & 215 \\
        Pitch & 16 & 268\\ 
        \bottomrule
    \end{NiceTabular}}
    \vspace{-5pt}
    \caption{
    We generate up to 200--500 short video clips with diverse scene configurations for parametric evaluation (\Cref{sec:App-PME}). Each video clip varies along one axis of distribution shift with a single target object. On average, each video has 300 frames.
    }
    \vspace{-15pt}
    \label{tab:param_eval_stat}
\end{table}


Parametric model evaluation is essential for developing and understanding perception models, enabling a systematic assessment of performance robustness against various domain shifts. Previous efforts, such as 3DCC~\cite{kar20223d}, have explored image corruption generation using 3D information, yet their scope is constrained by the static nature of input meshes, limiting the type and extent of possible variations. Leveraging the flexibility of the simulator, our generator extends parametric evaluation to more diverse axes, including scene, camera, and object state changes.

\vspace{2pt}
\noindent{\bf Task Design and Dataset Generation.} 
We focus on five key parameters difficult to rigorously control in real-world datasets yet significantly influence model performance: object articulation, lighting, object visibility, camera zoom, and camera pitch. Each parameter varies along a continuous axis for evaluating baseline models. For instance, object visibility varies from fully occluded to fully visible. 

\begin{figure}[t]
\begin{center}
\includegraphics[width=\linewidth]{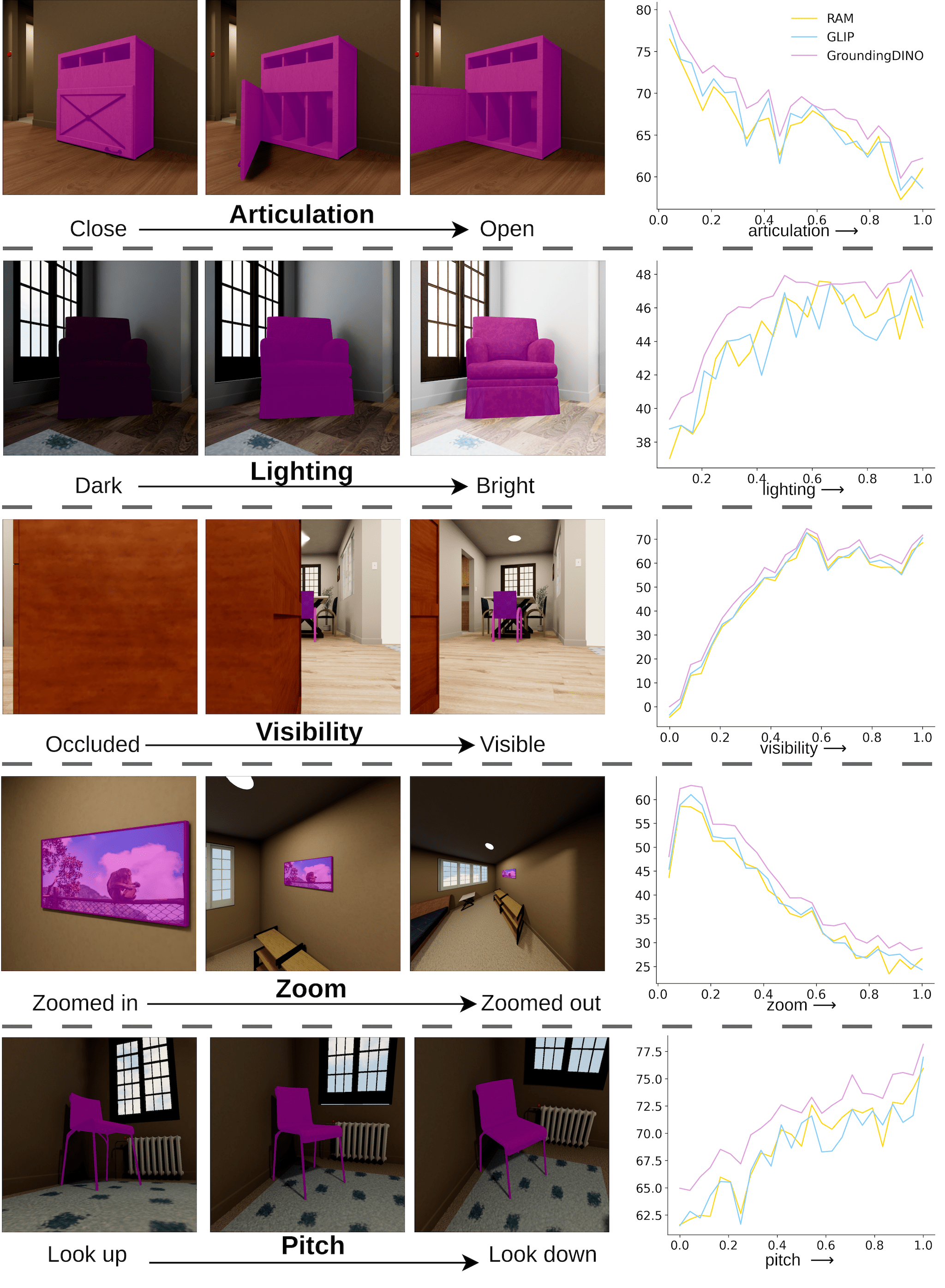}
\end{center}
\vspace{-20pt}
    \caption{Parametric evaluation of object detection models on five example video clips.
    Selected frames from these clips are shown on the left, with the target object highlighted in \textcolor{magenta}{magenta}.
    Average Precisions (APs) for our baseline models in \Cref{sec:App-HSU} are plotted on the right. Since \methodabbr allows for full customization of scene layout and camera viewpoints, we can systematically evaluate model robustness against variations in object articulation, lighting conditions, visibility, zoom (object proximity), and pitch (object pose). As illustrated, current SOTA models exhibit limited robustness to these axes of variation.
    }
\label{fig:parametric_dataset_and_plot}
\vspace{-10pt}
\end{figure}

We generate 200 to 500 videos for each axis (\Cref{tab:param_eval_stat}), using our collection of more than 8,000 3D assets. Each video includes a target object with changes focused on a single parameter under examination. \Cref{fig:parametric_dataset_and_plot} shows examples of target objects with variations along each axis.
We maintained consistency in other aspects of the environment, systematically synthesizing images to isolate the main parameter's impact.

To validate our findings in real-world conditions and further assess the sim2real transfer capability of parametric evaluation, we collected a smaller-scale real dataset for each of the 5 axes and replicated the evaluation. For setup details and additional results, please refer to the appendix.

\vspace{2pt}
\noindent{\bf Baselines and Metrics.} We explore two vision tasks: \textit{open-vocabulary detection} and \textit{open-vocabulary segmentation}, hypothesizing that models for these tasks may be sensitive to object-centric domain shifts. 
For baselines, we select the current state-of-the-art (SOTA) models on real datasets: GLIP \citep{li2022grounded}, RAM \citep{zhang2023recognize}, and Grounding DINO \citep{liu2023grounding} for detection, and ODISE \citep{zhang2023simple}, OpenSeeD \cite{xu2023open}, and Grounding SAM \cite{GroudingSAM} for segmentation.

\vspace{2pt}
\noindent{\bf Results and Analysis.}
In \Cref{fig:parametric_dataset_and_plot} and \Cref{fig:parametric_eval_radar}, we present example images when varying each parameter as well as respective detection Average Precision (AP) performance.
AP, calculated exclusively for the target object (highlighted in magenta), assesses the model's recognition accuracy. Detailed analyses reveal the following:

\begin{figure}[t]
    \centering
    \begin{subfigure}[t]{0.5\linewidth}
        \centering
        \includegraphics[width=0.98\linewidth]{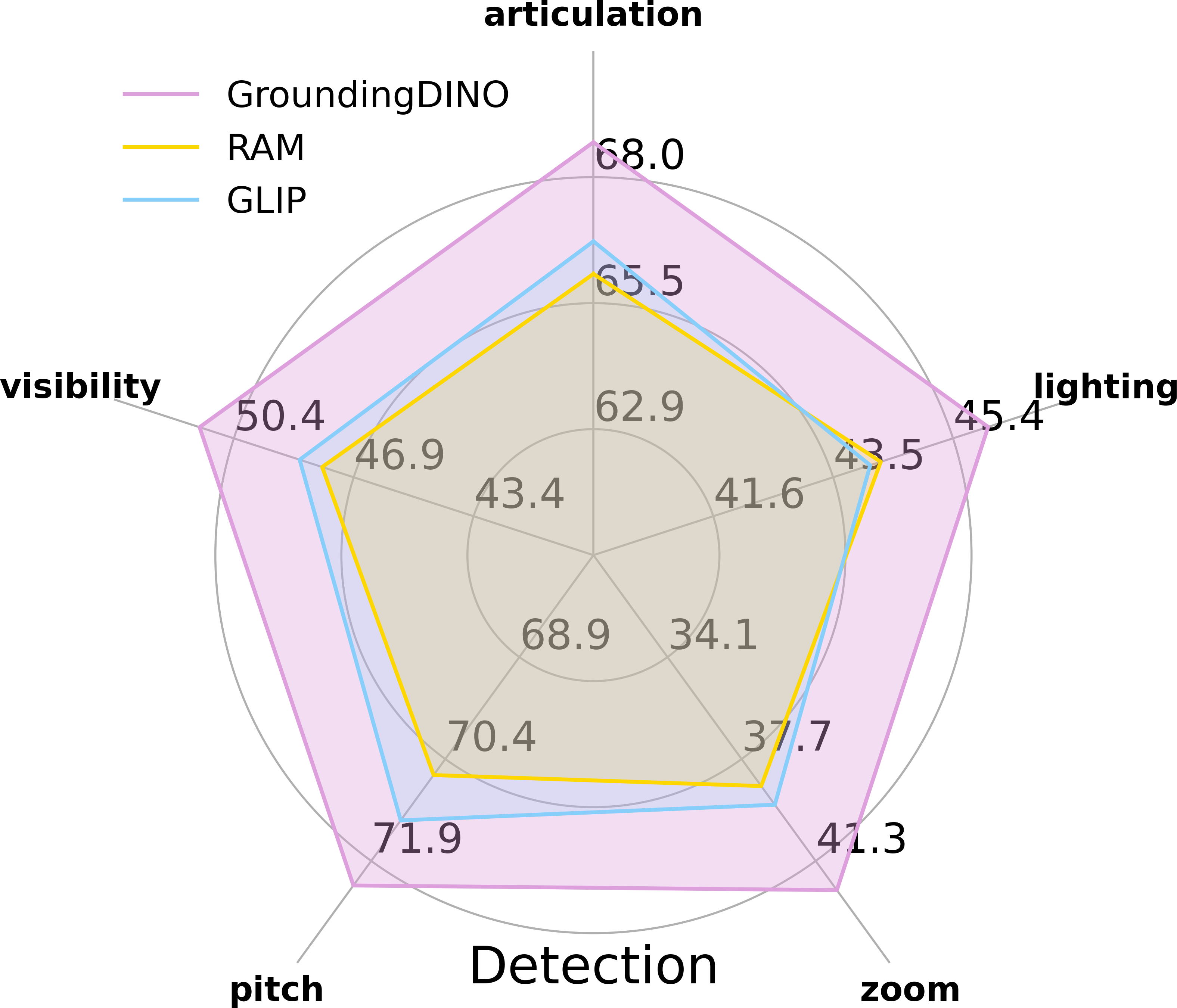}
    \end{subfigure}%
    \begin{subfigure}[t]{0.5\linewidth}
        \centering
        \includegraphics[width=0.98\textwidth]{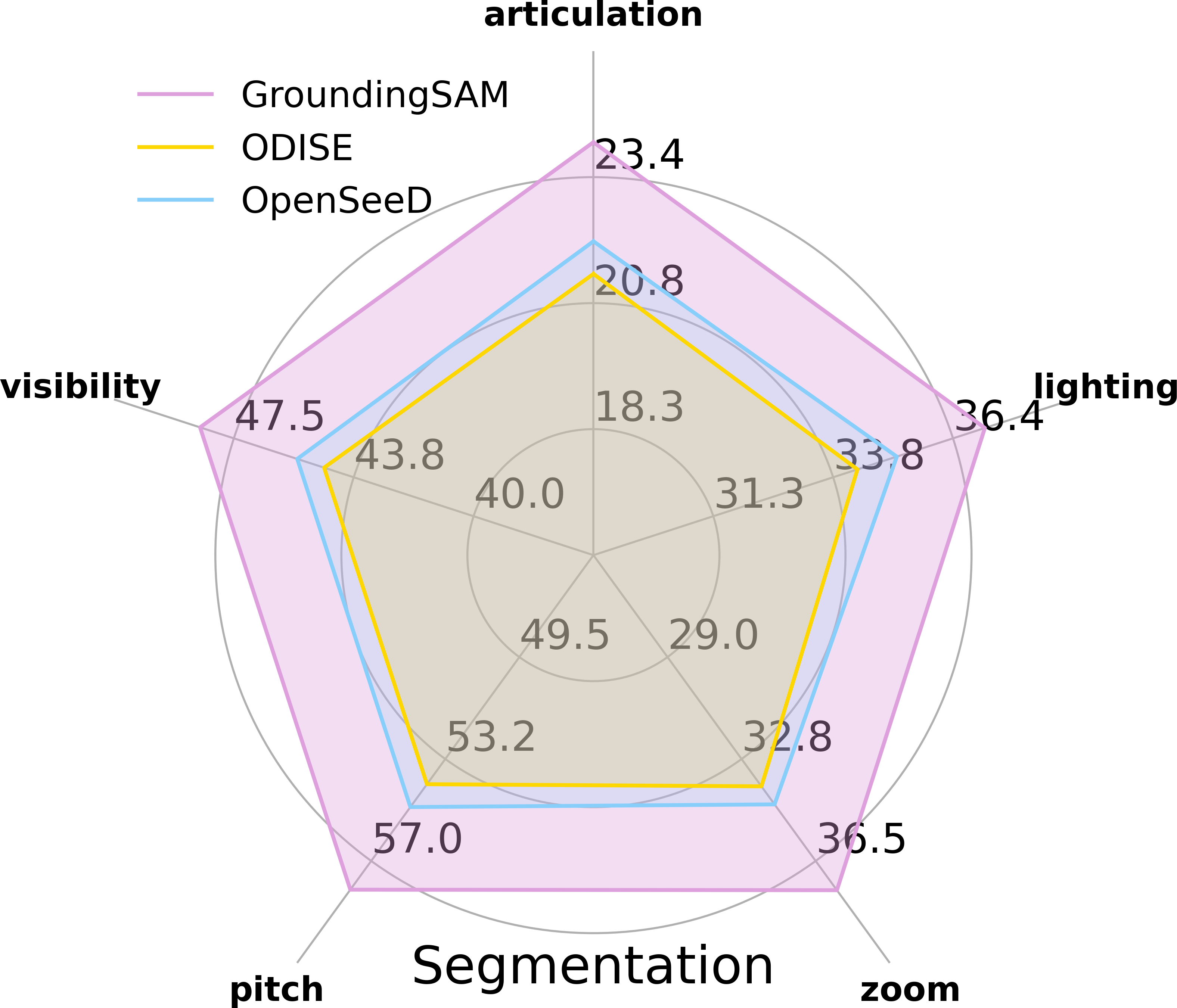}
    \end{subfigure}
    \caption{Mean performance of open-vocab object detection and segmentation models across five axes. 
    The larger a model's colored envelope, the more robust it is. Through \methodabbr, new vision models can be systematically tested for their robustness along these five dimensions and beyond: our users can easily add new axes of domain shift with just a few lines of code.}
\label{fig:parametric_eval_radar}
\vspace{-15pt}
\end{figure}

\begin{itemize}[nolistsep,topsep=1pt,leftmargin=1em,wide=\parindent]
    \item \textit{Articulation} varies the joint angles of the articulated target object, from fully closed to fully open, including processes such as opening/closing drawers or doors, and folding/unfolding laptops. 
A notable negative correlation between the degree of articulation and model performance suggests that models, typically trained or evaluated on existing benchmarks featuring mostly closed articulated objects (e.g., closed washing machines and microwaves), struggle with recognizing objects in open states. 
    
    \vspace{2pt}
    \item \textit{Lighting} adjusts global illumination of the environment from dark to bright.
We observed improving model performance up to a midpoint brightness level of 0.5, beyond which it plateaus. This suggests that, while current models suffer from low-light conditions, their performance saturates once the brightness level surpasses a certain threshold.

    \vspace{2pt}
    \item \textit{Visibility} shifts the visibility of the target object from fully occluded to fully visible, which is computed as the ratio of visible to total pixels of the target object.
We observe a steep decline in model performance as visibility drops below 0.5, which highlights a significant opportunity to enhance model robustness to partial occlusions.

    \vspace{2pt}
    \item \textit{Zoom} controls camera zoom from zoomed-in to zoomed-out. 
Results show that extremely close views, where a partial view of the target object occupies the entire image, hinder performance due to a lack of contextual information.
In contrast, too-zoomed-out views make the object too small for models to detect it effectively. Optimal performance is achieved at moderate zoom levels.


    \vspace{2pt}
    \item \textit{Pitch} varies camera pitch from looking up to looking down. 
We find that models perform inconsistently with seemingly benign changes in camera viewpoint, generally showing improved performance when the camera looks down at the target object. One potential explanation is that objects in large-scale real datasets are often captured from above, making this perspective more familiar to the models. 
\end{itemize}

\begin{figure*}[t]
\begin{center}
\includegraphics[width=0.8\linewidth]{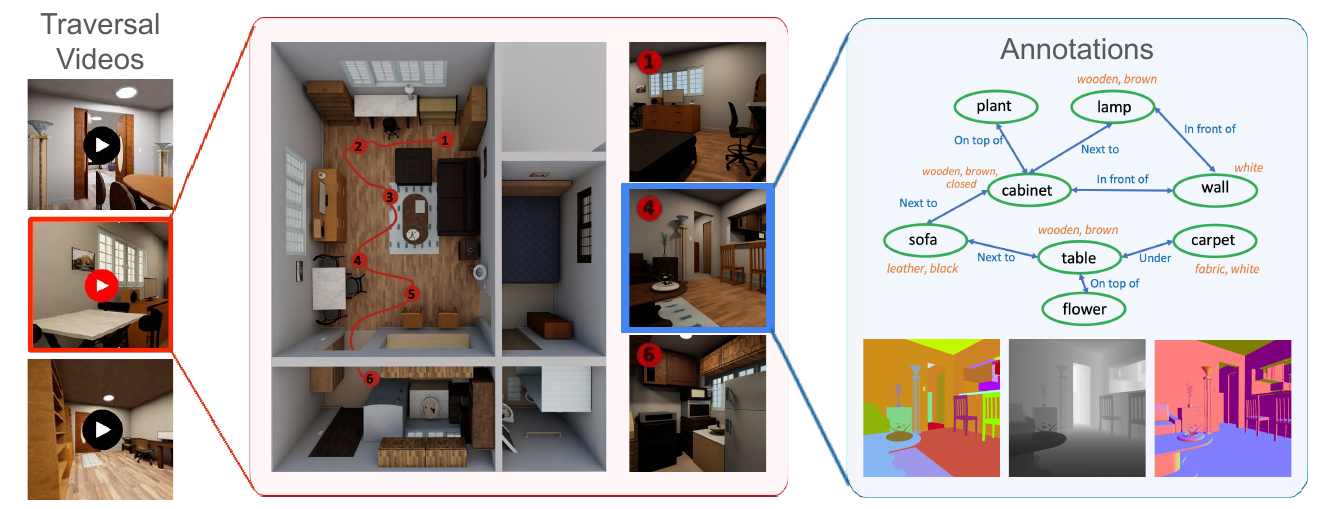}
\end{center}
\vspace{-20pt}
    \caption{Holistic Scene Understanding Dataset. We generated extensive traversal videos across representative scenes, each with 10+ camera trajectories. 
    For each image, \methodabbr generates various labels (e.g., scene graphs, segmentation masks, depth) as shown on the right. }
\label{fig:holistic}
\end{figure*}

\begin{table*}[t] 
    \begin{minipage}{\linewidth} 
        \begin{minipage}{0.49\linewidth}
            \centering
            \footnotesize
            \setlength{\tabcolsep}{2.62pt}
                \begin{tabular}{lcccc>{\columncolor[gray]{0.8}}c}
                    \toprule
                    \textbf{Open-vocab Det.} & AP $\uparrow$ & AP$_\text{\tiny small}$ $\uparrow$ & AP$_\text{\tiny medium}$ $\uparrow$ & AP$_\text{\tiny large}$ $\uparrow$ & COCO (AP) \\
                    \midrule
                    GLIP \cite{li2022grounded} & 41.4 & 7.0 & 27.5 & 61.8 & 60.8 \\
                    RAM \cite{zhang2023recognize} & 41.3 & 6.4 & 27.8 & 63.9 & 61.4 \\
                    Grounding DINO \cite{liu2023grounding} & \textbf{44.7} & \textbf{11.9} &  \textbf{31.2} & \textbf{66.3} & 63.0 \\
                    \bottomrule
                \end{tabular}
            \label{tab:open-vocab-detection}
        \end{minipage}\hfill
        \begin{minipage}{0.49\linewidth}
            \centering
            \footnotesize
            \setlength{\tabcolsep}{2.62pt}
                \begin{tabular}{lcccc>{\columncolor[gray]{0.8}}c}
                    \toprule
                    \textbf{Open-vocab Seg.} & AP $\uparrow$ & AP$_\text{\tiny small}$ $\uparrow$ & AP$_\text{\tiny medium}$ $\uparrow$ & AP$_\text{\tiny large}$ $\uparrow$ & ADE (AP) \\
                    \midrule
                    ODISE \cite{xu2023open} & 57.1 & 41.0 & 53.2 & 65.0 & 13.9 \\
                    OpenSeeD \cite{zhang2023simple} & 57.3 & 42.0 & 54.1 & 64.8 & 15.0 \\
                    Grounding SAM \cite{GroudingSAM} & \textbf{59.2} & \textbf{42.9} & \textbf{54.4} & \textbf{65.1} & 14.8 \\
                    \bottomrule
                \end{tabular}
            \label{tab:open-vocab-segmentation}
        \end{minipage}
    \end{minipage}
    
    \begin{minipage}{\linewidth}
        \begin{minipage}{0.49\linewidth}
            \centering
            \footnotesize
            \setlength{\tabcolsep}{3pt}
                \begin{tabular}{lcccccc>{\columncolor[gray]{0.8}}c}
                    \toprule
                    \textbf{Depth Est.} & RMS $\downarrow$ & AbsRel $\downarrow$ & Log10 $\downarrow$ & $\delta_1$ $\uparrow$ & $\delta_2$ $\uparrow$ & $\delta_3$ $\uparrow$ & NYUv2 ($\delta_1$) \\
                    \midrule
                    DPT \cite{Ranftl2021} & 0.66 & 0.14 & 0.05 & 0.09 & 0.15 & 0.20 & 0.90 \\
                    NVDS \cite{wang2023neural} & 0.58 & \textbf{0.13} & \textbf{0.04} & 0.10 & 0.15 & 0.21 & 0.93 \\
                    iDisc \cite{piccinelli2023idisc} & \textbf{0.49} & \textbf{0.13} & \textbf{0.04} & \textbf{0.12} & \textbf{0.19} & \textbf{0.22} & 0.94 \\
                    \bottomrule
                \end{tabular}
            \label{tab:depth-estimation}
        \end{minipage}\hfill
        \begin{minipage}{0.49\linewidth}
            \centering
            \footnotesize
            \setlength{\tabcolsep}{3pt}
                \begin{tabular}{lccc>{\columncolor[gray]{0.8}}c}
                    \toprule
                    \textbf{Point Cloud Recon.} & Comp. Ratio $\uparrow$ & Comp. $\uparrow$ & Acc. $\downarrow$ & Replica (C.R.) \\
                    \midrule
                    GradSLAM \cite{jatavallabhula2019gradslam} & 50.0 & \textbf{14.8} & 29.8 & 67.9 \\
                    NICE-SLAM \cite{zhu2022nice} & \textbf{66.3} & 12.0 & \textbf{23.5} & 89.3 \\
                    \bottomrule
                \end{tabular}
            \label{tab:3d-scene-reconstruction}
        \end{minipage}
    \end{minipage}
    \caption{
    A comprehensive evaluation of SOTA models on four vision tasks.
    Our synthetic dataset can be a faithful proxy for real datasets as the relative performance between different models closely correlates to that of the \colorbox{gray!30}{real datasets}.
    }
    \label{fig:holistic_result}

\end{table*}

To summarize, we observe significant performance discrepancies across three models on all five axes, with our parallel experiments in real settings (see the appendix) confirming that these trends observed in synthetic data mirror those in real-world scenarios. This underscores the lack of robustness of the current SOTA models in extreme or out-of-distribution test environments. By generating large-scale synthetic datasets with controlled variability, \methodabbr provides a unique and powerful test bed to evaluate model performance. 
Furthermore, consistent with \Cref{sec:App-HSU}, the relative performance remains steady across the five axes, highlighting the predictive value of the datasets generated by \methodabbr.


\subsection{Holistic Scene Understanding}
\label{sec:App-HSU}

One of the major advantages of synthetic datasets, including \methodabbr, is that they offer various types of labels (segmentation masks, depth maps, and bounding boxes) for the same sets of input images. We believe that this feature can fuel the development of versatile vision models that can perform multiple perception tasks at the same time in the future. Since such models are not currently available, we instead evaluate the current SOTA methods on a subset of the tasks that \methodabbr supports (see below). This will also serve as a validation of the photorealism of our datasets, i.e., models trained on real datasets should perform reasonably without fine-tuning. 

\vspace{2pt}
\noindent{\bf Task Design and Dataset Generation.} Equipped with \methodabbr's powerful generator (see~\Cref{sec:BEHAVIOR-Dataset-Generator}), we generated 100+ full scene traversal videos with a total of 266240 frames with per-frame ground truth annotations in multiple modalities. \Cref{fig:holistic} shows an overview of the generated dataset.



\vspace{2pt}\noindent{\bf Baselines and Metrics.} In \Cref{fig:holistic_result}, we assess 11 models in four tasks. Specifically, we consider \textit{Detection} and \textit{Segmentation} tasks, both in the challenging open vocabulary setting~\cite{zhang2023simple, liu2023grounding}, as well as \textit{Depth Estimation} and \textit{Point Cloud Reconstruction}, with standard metrics used. 

\vspace{2pt}
\noindent{\bf Results and Analysis.}  We summarize all our evaluation results in~\Cref{fig:holistic_result}. We observe that the relative performance of these models on our synthetic dataset has a high correlation with that on real datasets such as MS COCO~\cite{cocodataset} or NYUv2~\cite{NYUv2}, indicating that our generated synthetic datasets can be a faithful proxy for real datasets.

In summary, we provide a comprehensive benchmark to score and understand a wide range of existing models for each of the four tasks on exactly the same images.
Although most current vision models focus on a single output modality, we hope \method could motivate researchers and practitioners to develop versatile models that concurrently predict multiple modalities in the future, where our benchmarking results for single-task SOTA methods in this section could serve as a useful reference.

\subsection{Object States and Relations Prediction}
\label{sec:App-UB}

    



\begin{figure}[t]
\begin{center}
\includegraphics[width=0.8\linewidth]{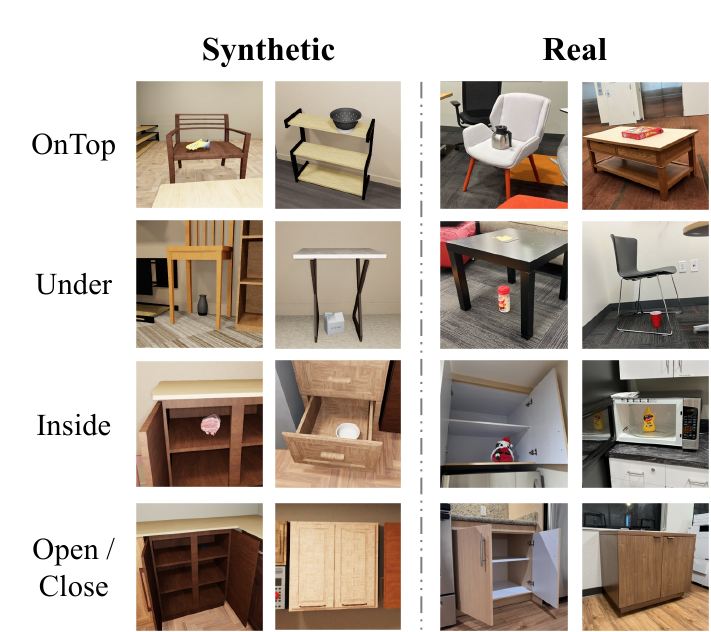}
\end{center}
\vspace{-15pt}
    \caption{Sample images of each class from our generated synthetic and collected real datasets.}
\vspace{-5pt}
\label{fig:spatial}
\end{figure}

\methodabbr's capabilities extend beyond model evaluation shown in \Cref{sec:App-PME} and \Cref{sec:App-HSU}. Users can also leverage \methodabbr to generate training data with specific object configurations that are difficult to accumulate or annotate in the real world. This section illustrates \methodabbr's practical application in synthesizing a dataset that facilitates the training of a vision model capable of zero-shot transfer to real-world images on the task of object relationship prediction~\cite{he2022synthetic, ge2022neural, ge2022empaste, ge2023beyond, ge2022dall, cascante2023going}. Additional experiments focusing on unary object states, \texttt{filled} and \texttt{folded}, are detailed in the appendix.

\vspace{2pt}
\noindent{\bf Task Design and Dataset Generation.} Predicting object relationships, such as \texttt{open} and \texttt{inside}, is a crucial yet challenging perception task due to the difficulties in collecting such data in the real world, let alone the costly annotations\cite{krishna2017visual, mao2018neuro, grounding_predicates}. 
We use our generator to synthesize 12.5k images with five labels (\texttt{open}, \texttt{close}, \texttt{ontop}, \texttt{inside}, \texttt{under}), depicting relationships between target objects. We also collected and labeled 910 real images with unseen object instances and scenes to test sim2real performance. Examples are shown in \Cref{fig:spatial}. 

\vspace{2pt}
\noindent{\bf Baselines and Metrics.}
Adapting from~\cite{grounding_predicates}, our model takes an image and target objects' bounding boxes as input, and outputs a five-way classification over the five labels. 
We specifically define \texttt{open/close} as a binary relationship between the movable link and the unmovable base of an articulated object, enabling fine-grained articulation state assessment. For example, the model can be queried for the open or closed status of individual drawers of a cabinet. 
Detailed model architecture is available in the appendix.

We compare our model with zero-shot CLIP, which is not trained on our synthetic dataset, in terms of precision, recall, and F1, on the synthetic evaluation set and the real test set. Specifically, by harnessing CLIP's zero-shot capabilities~\cite{radford2021learning}, this baseline outputs a five-way classification prediction by comparing the image embeddings with the five verbalized prompts' text embeddings.



\begin{table}[t]
    \small
    \centering
    \setlength{\tabcolsep}{3pt}
    \begin{NiceTabular}[baseline=2,cell-space-limits=1pt]{lccccccc} \toprule
        \Block{1-2}{Test on} & & Open & Close & Ontop & Inside & Under & Avg. \\
        \midrule
        \Block{2-1}{Synthetic} & Precision &  0.962 &  0.897&  0.947&  0.989& 0.874 & 0.932 \\
        & Recall &  0.822 & 0.978 & 0.913 & 0.995 & 0.949 & 0.929  \\
        \cmidrule{2-8}
         \Block{2-1}{Real} & Precision & 0.943 & 0.958 & 0.545 & 0.906 & 0.948 & 0.863 \\
        & Recall & 0.757 & 0.915  & 0.913 & 0.776 & 0.703 & 0.817 \\
        \bottomrule
    \end{NiceTabular}
    \caption{
    Classification results on held-out synthetic eval set and real test set for our method adapted from~\cite{grounding_predicates}. 
    }
    \label{tab:spatial_stat}
\end{table}

\begin{table}[t]
    \small
    \centering
    \begin{NiceTabular}[baseline=2,cell-space-limits=1pt]{lccc} 
        \toprule
        Method & Precision & Recall & F1 \\ 
        \midrule
        Zero-shot CLIP & 0.293 & 0.282 & 0.271 \\
        Ours & \textbf{0.863} & \textbf{0.817} & \textbf{0.839} \\
        \bottomrule
    \end{NiceTabular}
    \caption{
    Classification results on the real test set. 
    Task-specific training on synthetic data boosts performance on real images.
    }
    \label{tab:pred_result_comp}
\end{table}

\vspace{2pt}
\noindent{\bf Results and Analysis.} 
\Cref{tab:spatial_stat} presents the quantitative results on the held-out synthetic dataset and the real dataset for our method. Although there is some performance gap, our model trained on only synthetic data can transfer to real images with promising overall accuracy. Additionally, unary state prediction experiments, detailed in the appendix, also reveal high accuracy in both domains.
These results underscore that \methodabbr offers a promising way to obtain realistic synthetic data that researchers can use not only for evaluation (as shown in \Cref{sec:App-PME} and \Cref{sec:App-HSU}), but also for training models that can then be transferred to the real world. In fact, from~\Cref{tab:pred_result_comp}, we observe that task-specific training on synthetic data is a very effective method to obtain good performance on real images. 

\section{Conclusion}
\label{sec:conclude}
We have introduced the \method (\methodabbr), a novel toolkit designed for the systematic evaluation and comprehensive understanding of computer vision models. \methodabbr enables researchers to control a wide range of parameters across scene, object, and camera levels, facilitating the creation of highly customized datasets.
Our experiments highlight \methodabbr's versatility and efficacy through three key applications. First, we show its ability to evaluate model robustness against various domain shifts, underscoring its value in systematically assessing model performance under challenging conditions. Second, we present comprehensive benchmarking of scene understanding models on a unified dataset, illustrating the potential for developing multi-task models using a single \methodabbr dataset. Lastly, we investigate \methodabbr's role in facilitating sim2real transfer for novel vision tasks, including object states and relations prediction. 
\methodabbr highlights synthetic data's promise in advancing the field, offering researchers the means to generate high-quality, diverse, and realistic datasets tailored to specific needs. 




\paragraph{Acknowledgments.}
We are grateful to SVL members for their helpful feedback and insightful discussions.
The work is in part supported by the Stanford Institute for Human-Centered AI (HAI), NSF CCRI \#2120095, RI \#2338203, ONR MURI N00014-22-1-2740, N00014-21-1-2801, Amazon, Amazon ML Fellowship, and Nvidia.

\clearpage

{\small
    \bibliographystyle{ieeenat_fullname}
\bibliography{0-main.bbl}
}
\clearpage
\appendix
\maketitlesupplementary

\section{Extended BEHAVIOR-1K Assets}

\begin{figure*}[t]
\begin{center}
\includegraphics[width=0.65\textwidth]{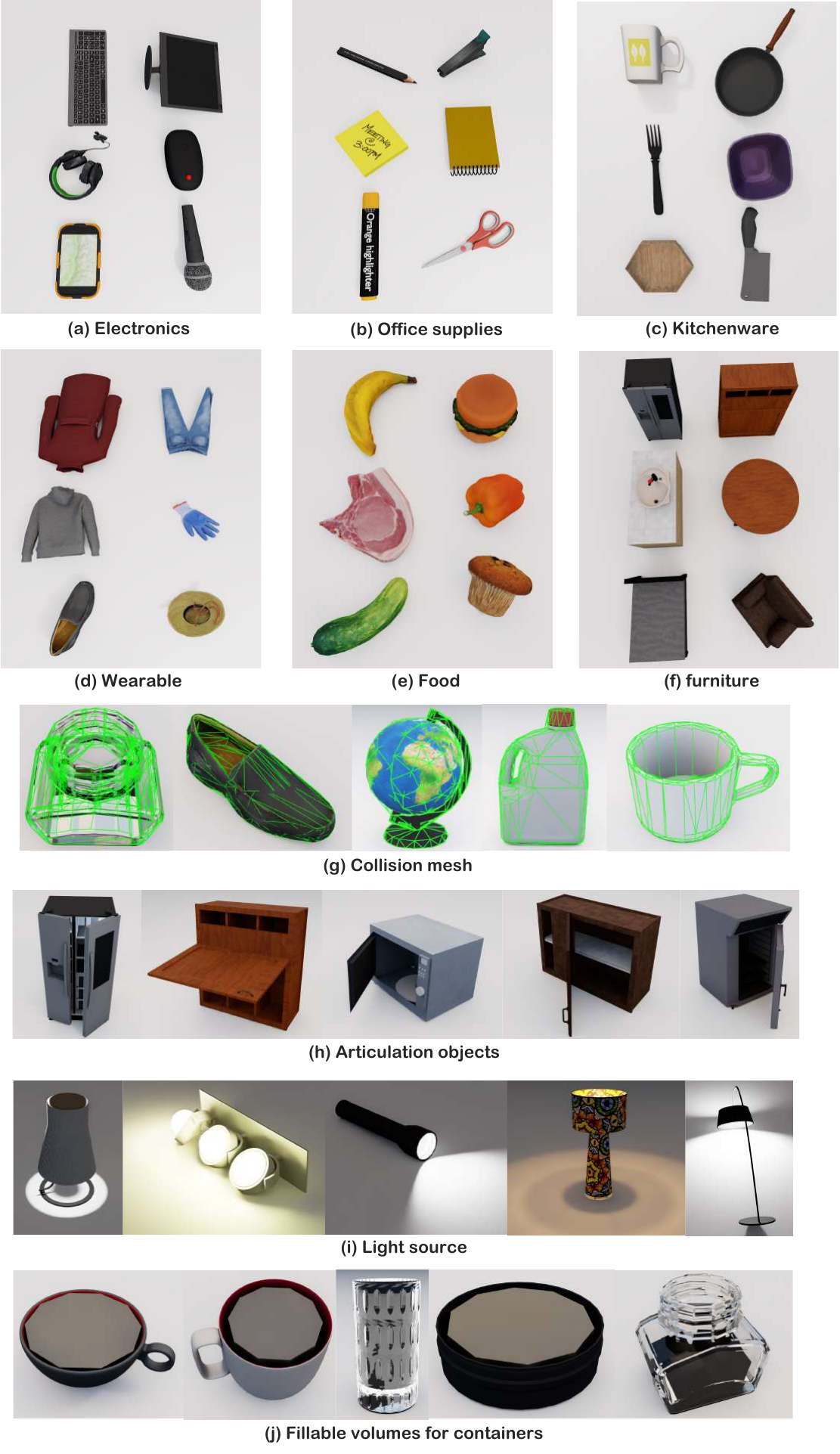}
\end{center}
    \caption{More examples of Extended BEHAVIOR-1K Assets: (a-f) examples for different main categories. (g) examples of improved collision mesh quality. (h) examples of articulation objects. (i) examples of different light sources, (j) examples of fillable volumes for containers.}
\label{fig:supp-assets}
\end{figure*}

Fig.~\ref{fig:supp-assets} shows more examples of Extended BEHAVIOR-1K Assets (main paper Sec.3.1): (a-f) examples for different main categories. (g) examples of improved collision mesh quality. (h) examples of articulation objects. (i) examples of different light sources, (j) examples of fillable volumes for containers.

\section{Customizable Dataset Generator}
We prepare a \textbf{video} in the supplementary material to show an example visualization of the Customizable Dataset Generator: specifically, we show the scene instance augmentation by scene object (furniture) randomization and inserting additional everyday objects. 




\section{Experiments}

\subsection{Parametric Model Evaluation}

\paragraph{Details of Dataset Generation Process.}
We synthesized the evaluation videos for each axis (Object articulation, Lighting, Visibility, Zoom, Pitch) according to the following pipeline. 

As shown in the main paper Sec. 4.1, each video includes a target object with changes focused on a single parameter under examination.
First, we sample one of the scene instances and randomly choose a target object in the scene. For the object articulation axis, we only sample objects with movable parts, such as cabinets, microwaves, refrigerators, etc. Next, we sample a random camera angle and distance with the target object placed in the center. 
Then, for all except pitch, we keep this camera pose and perform an axis-specific manipulation to generate a video with the desired variation:
\begin{itemize}
    \item \textbf{Object articulation}: We linearly interpolate the joint angle from being closed to fully open, utilizing the joint maximum range annotations provided in \methodabbr assets. We record the image with the joint in each intermediate state. For objects with multiple movable parts, e.g., a cabinet with three drawers, we randomly sample a subset of joints to manipulate and keep the rest closed. 
    \item \textbf{Lighting}: We linearly increase the intensity of all indoor light sources in the scene simultaneously.
    \item \textbf{Visibility}: There are three key components in the visibility (occlusion) setting: camera, target object, and occluding object. We first set the camera centering on the target object, then we place an occluding object (relatively large object, e.g., cabinet) in the line between the camera and the target object, fully occluding the target in view. Then, we fix the distance between the camera and the target object and move the camera around the target object until the target is fully visible. 
    The visibility score (number of visible pixels/number of total pixels) of each frame is calculated by rendering the video again and removing the occluding cabinet. Although the object orientation in camera view might slightly change since the camera is not static, we implemented the following practices to eliminate the effect of this factor. First, we set the camera relatively far from the target but occluding objects close to the camera, allowing minimal camera pose change needed to capture the "fully occluded to fully visible" process. In addition, the initial object pose is randomized, so when we average evaluation performance, the effect of this factor shall largely cancel out. 
    \item \textbf{Zoom}: With the camera pose fixed, we change its focal length to model the zooming effect. We strategically changed the focal length such that the resulting video illustrates an approximately linear zooming behavior. We always make the target object at the center of the view, and it remains mostly unaffected by distortion, even under extreme focal length.
    \item \textbf{Pitch}: We linearly change the camera pitch angle while keeping the original camera distance as well as the yaw angle unmodified.
\end{itemize}
After each video was collected, we filtered out the videos where the target object was not properly visible. 
The detailed statistics for each axis is shown in main Table 2.

\paragraph{Metric Details.}
Each generated video contains exactly one target object (main paper Figure 3 \textcolor{magenta}{magenta}).
We use different open-vocabulary object detection and segmentation models to detect or segment the target object. These models act as indicators of performance in challenging environments, such as those with limited lighting or long-distance zoom.
Therefore, we compute the Average Precision (AP) metric using the target object as the sole ground truth, considering only predictions that classify the target object class.
However, it is plausible to encounter scenarios where multiple objects of the same category as the target object exist. For instance, in a video, there might be several chairs, and the target object is one of those chairs.
Models might have correctly detected all chairs, but since only one is the ground truth, all rest will be marked as incorrect.
To counter such an undesired situation, we employ a simple heuristic to filter predictions for the non-target object:
For non-target objects sharing the same category, we calculate the IoU with each prediction and exclude those with an IoU exceeding a predefined threshold of 0.3. This means treating these predictions as valid for non-target objects rather than false positives.
This threshold is chosen empirically based on a few selected cases where objects of the same category are densely packed together. We believe this choice generalizes well to less crowded scenes and ensures the reliability of our evaluation process.

\begin{figure*}
    \centering
    \begin{subfigure}[b]{\linewidth}
        \includegraphics[width=\linewidth]{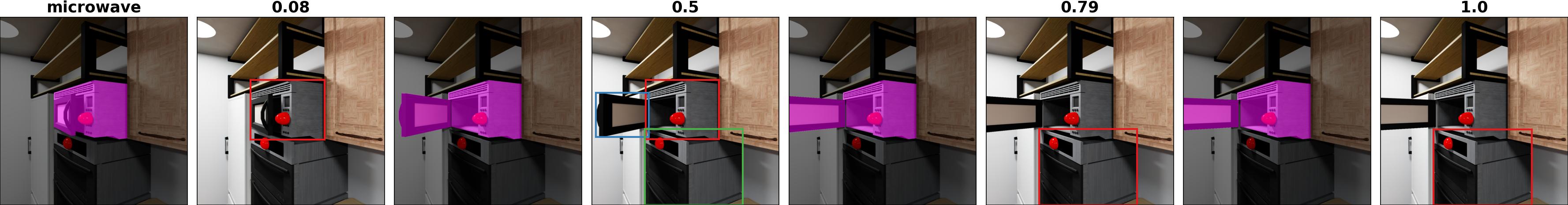}
        \caption{
       \textbf{Articulation.} The model may have limited exposure to open states for articulation objects, which makes it less likely to recognize a microwave with its door open correctly.
        }
    \end{subfigure}%
    \hfill
    \begin{subfigure}[b]{\linewidth}
        \includegraphics[width=\linewidth]{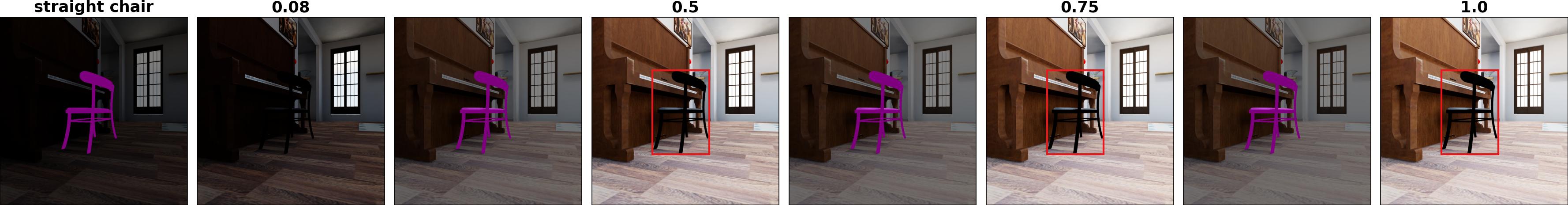}
         \caption{
       \textbf{Lighting.} When the environment is dark, the model performance is negatively affected. However, when the lighting exceeds a certain threshold, in this case 0.5, the model becomes robust to increasing illumination.
        }
    \end{subfigure}%
    \hfill
    \begin{subfigure}[b]{\linewidth}
        \includegraphics[width=\linewidth]{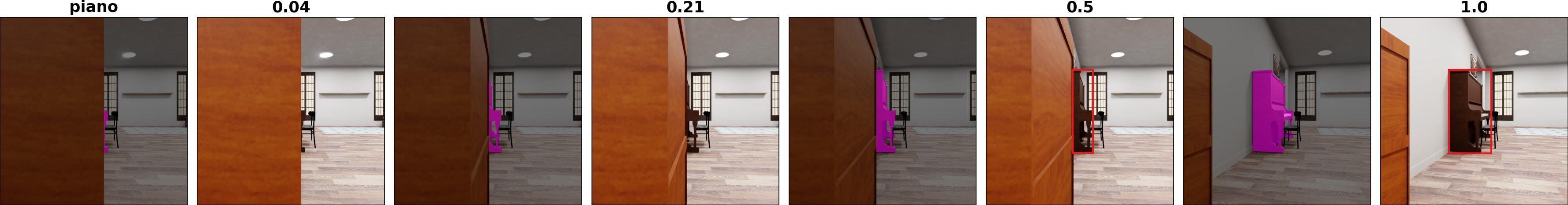}
         \caption{
        \textbf{Visibility.} The model's detection performance suffers when most of the target object is occluded. Surprisingly, a correct prediction can be made with only half of the object visible.
        }
    \end{subfigure}%
    \hfill
    \begin{subfigure}[b]{\linewidth}
        \includegraphics[width=\linewidth]{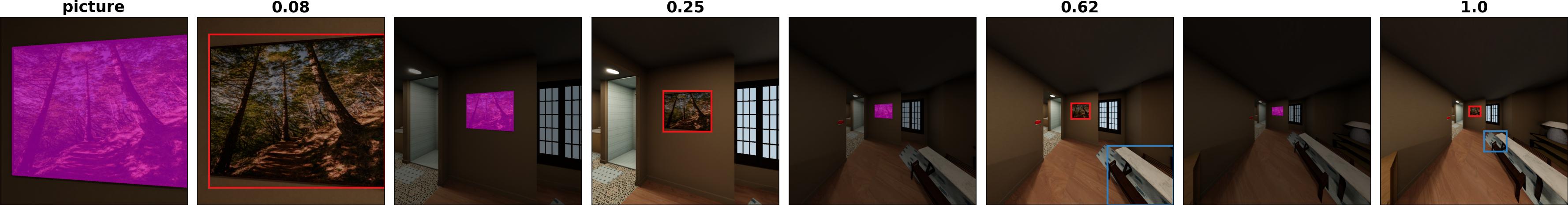}
         \caption{
         \textbf{Zoom.} When the model is zoomed out, nearby objects become distorted, leading the model to identify irrelevant objects as the target mistakenly. This suggests that the model's recognition may rely partly on contours rather than solely on semantic information.
        }
    \end{subfigure}%
    \hfill
    \begin{subfigure}[b]{\linewidth}
        \includegraphics[width=\linewidth]{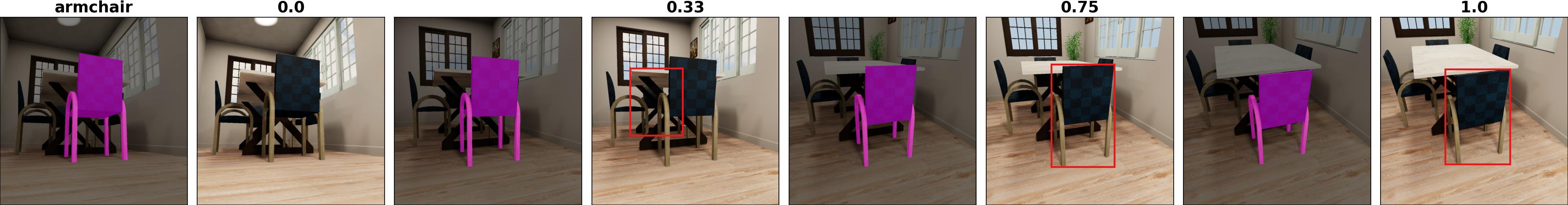}
         \caption{
         \textbf{Pitch.} We find that, generally, the model can achieve better performance in a look-down angle compared to a look-up angle.
        }
    \end{subfigure}%
    \caption{Error analysis for Grounding DINO \cite{liu2023grounding}. Similar trends are also observed in other detection models. Each row in our presentation represents one axis and comprises four example groups. In each group, there are two images: the left image illustrates the ground truth, highlighting the target object in \textcolor{magenta}{magenta}, while the right image shows the Grounding DINO's predictions (colored differently) for the target object, as indicated at the top of the first image in each group. 
The example groups are arranged such that, from left to right, the intensity along the respective axis increases (e.g., progressing from zoomed in to zoomed out), and the intensity value (0-1) is shown on top of each prediction.}
    \label{fig:error_analysis_detection}
\end{figure*}

\paragraph{Failure Case Analysis on Five Evaluation Parameters.}
In \Cref{fig:error_analysis_detection}, we present failure case examples of Grounding DINO \cite{liu2023grounding} across five evaluation axes: Object articulation, Lighting, Visibility, Zoom, and Pitch.
Each row in our presentation represents one axis and comprises four example groups. In each group, there are two images: the left image illustrates the ground truth, highlighting the target object in \textcolor{magenta}{magenta}, while the right image shows the Grounding DINO's prediction for the target object, as indicated at the top of the first image in each group. 
The example groups are arranged such that, from left to right, the intensity along the respective axis increases (e.g., progressing from zoomed in to zoomed out), the intensity value (0-1) is shown on top of each prediction.
We find and highlight some interesting findings for each parametric evaluation in \Cref{fig:error_analysis_detection} and detailed below.
For the full axis, we prepare a \textbf{video} in supplementary to show qualitative examples about running different detection models on the same video. 
\begin{itemize}
    \item \textbf{Articulation.} The model may have limited exposure to open states for articulation objects, which makes it less likely to predict a microwave with its door open correctly.
    \item \textbf{Lighting.} When the environment is dark, the model performance is negatively affected. However, when the lighting exceeds a certain threshold, in this case 0.5, the model becomes robust to increasing illumination.
    \item \textbf{Visibility.} The model's detection performance suffers when most of the target object is occluded. Surprisingly, a correct prediction can be made with only half of the object visible.
    \item \textbf{Zoom.} When the model is zoomed out, nearby objects become distorted, leading the model to identify irrelevant objects as the target mistakenly. This suggests that the model's recognition may rely partly on contours rather than solely on semantic information.
    \item \textbf{Pitch.} We find that, generally, the model can achieve better performance in a look-down angle compared to a look-up angle.
\end{itemize}

\begin{figure*}
    \centering
    \begin{subfigure}[b]{0.45\linewidth}
        \includegraphics[width=\linewidth]{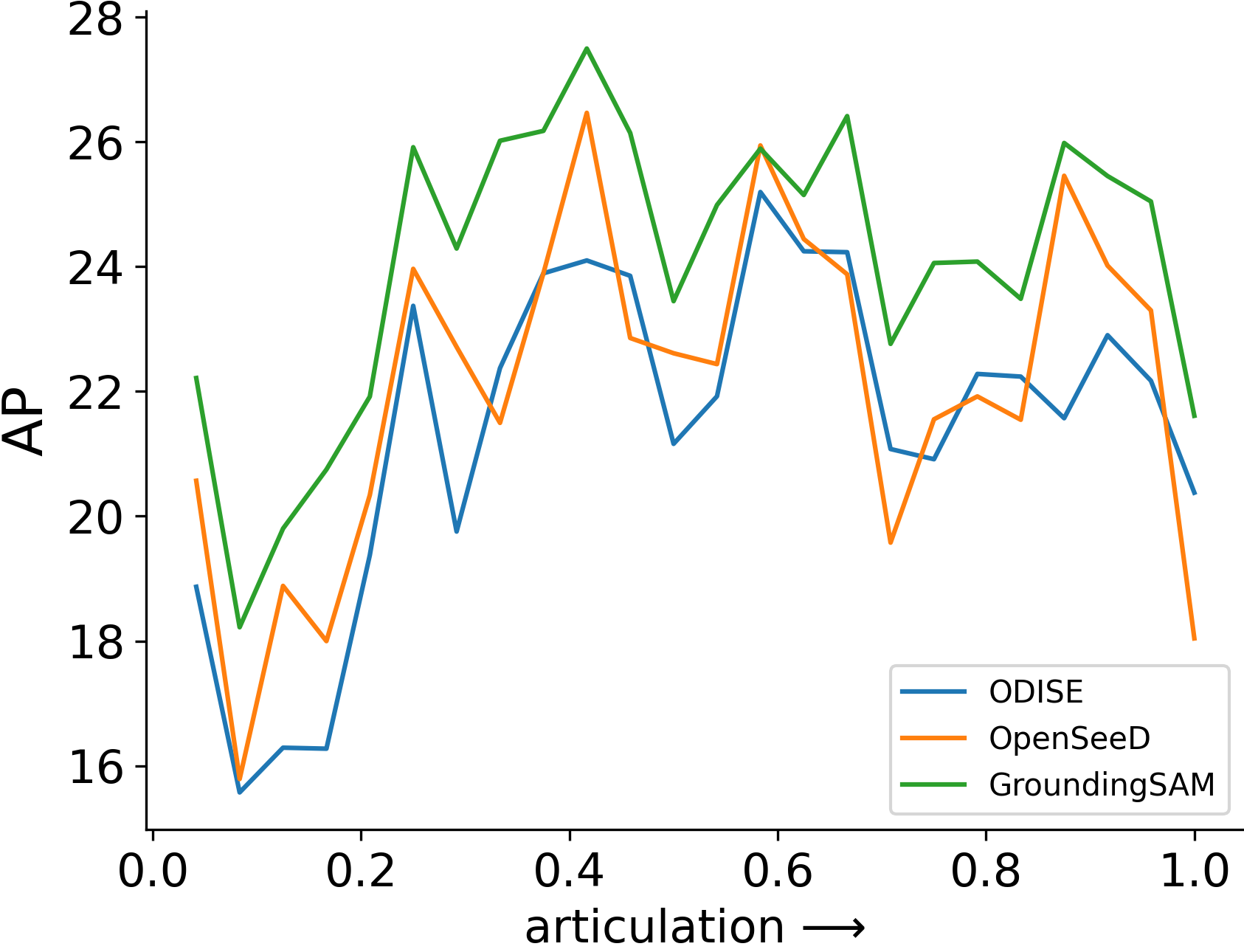}
    \end{subfigure}%
    \hfill
    \begin{subfigure}[b]{0.45\linewidth}
        \includegraphics[width=\linewidth]{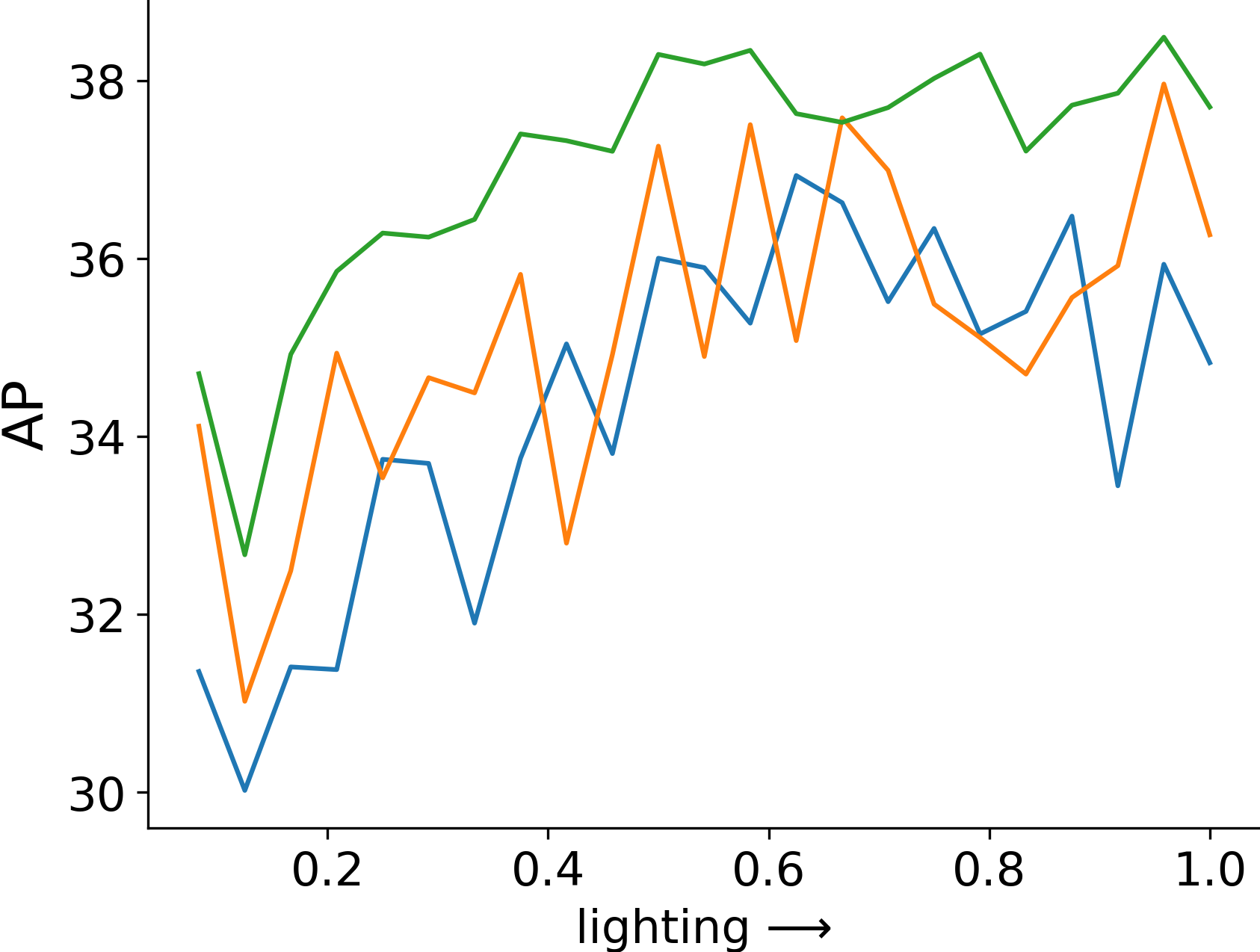}
    \end{subfigure}%
    \hfill
    \begin{subfigure}[b]{0.45\linewidth}
        \includegraphics[width=\linewidth]{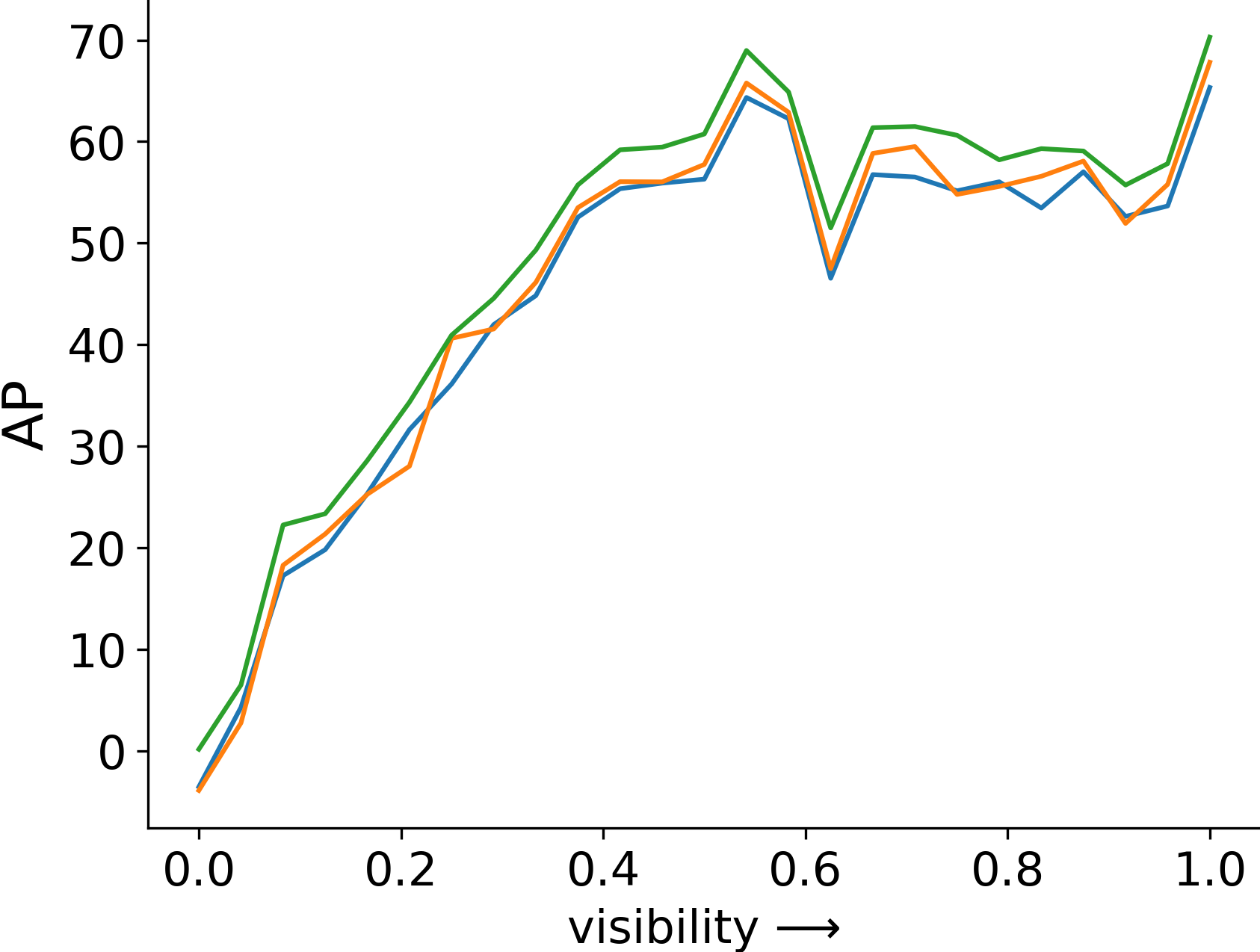}
    \end{subfigure}%
    \hfill
    \begin{subfigure}[b]{0.45\linewidth}
        \includegraphics[width=\linewidth]{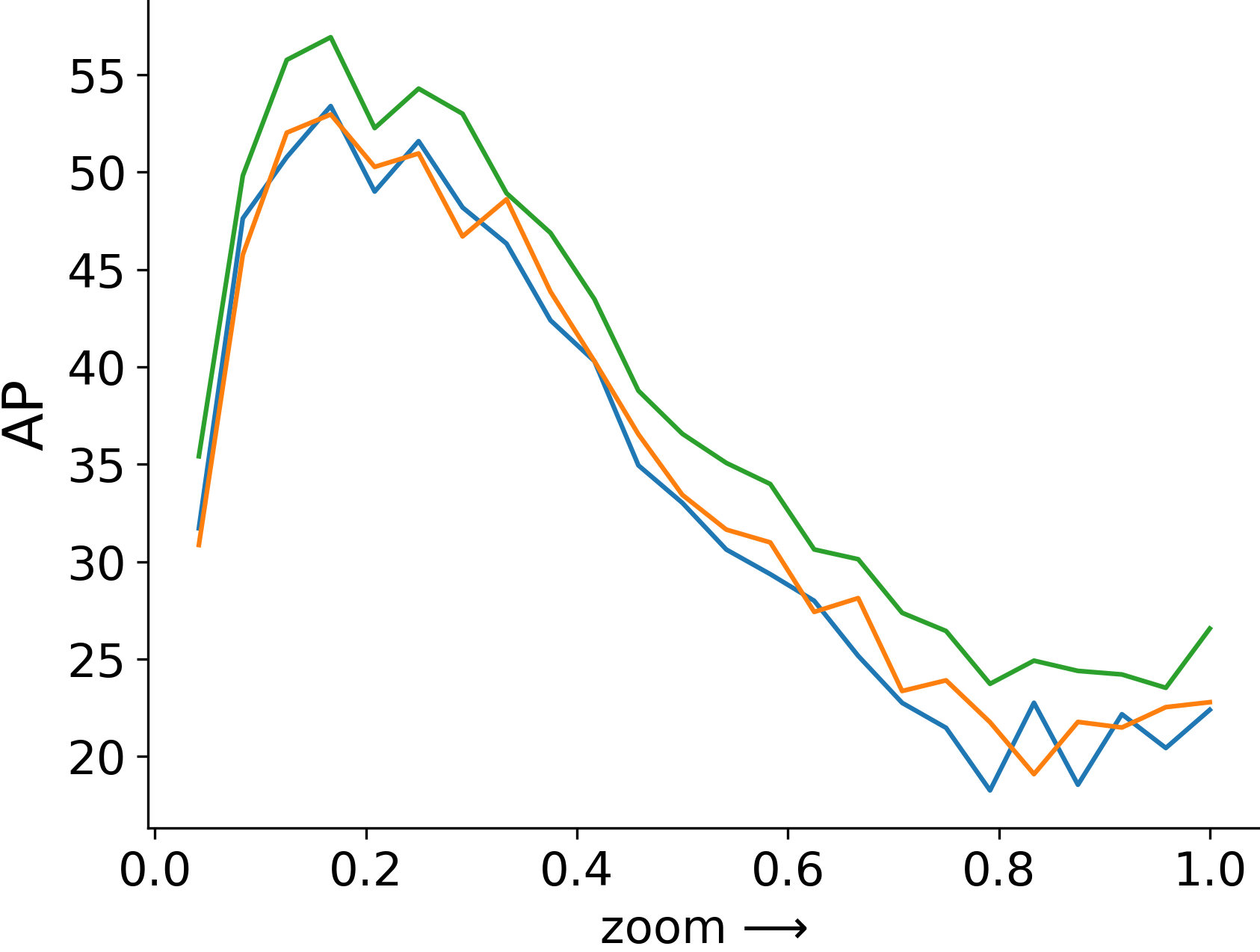}
    \end{subfigure}%
    \hfill
    \begin{subfigure}[b]{0.45\linewidth}
        \includegraphics[width=\linewidth]{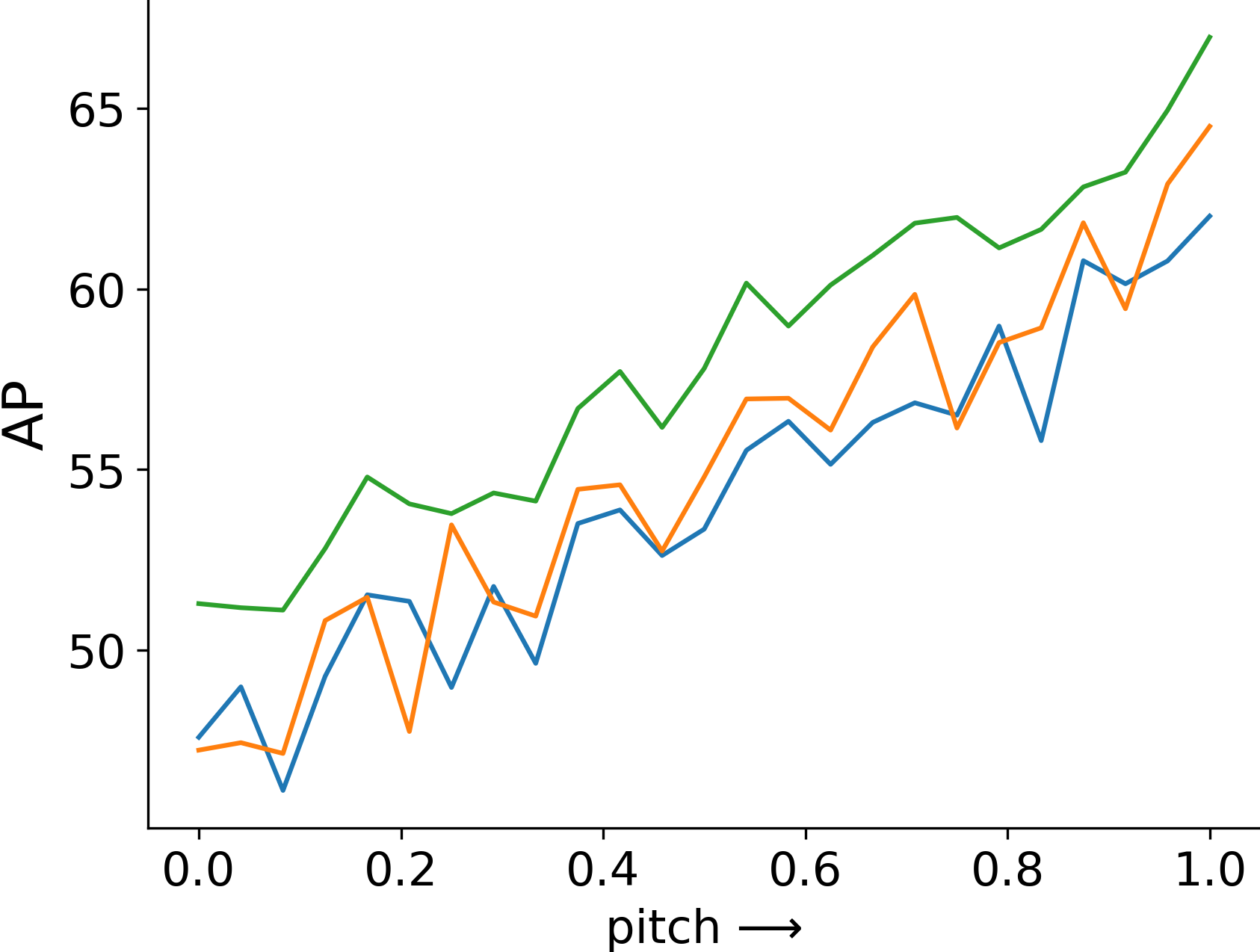}
    \end{subfigure}%
    \caption{Similar to Figure 3 in the main (Section 4.1), we plot the Segmentation model results for each of the five axes. AP is calculated using target object only.}
    \label{fig:seg_param_eval_plots}
\end{figure*}

\paragraph{Segmentation Results on Five Axes.}
Figure 3 of the main paper shows the performance of open-vocabulary detection models on five axes. In \Cref{fig:seg_param_eval_plots}, we show the performance of open-vocabulary segmentation models instead.
    The average performance for each axes corresponds to one angle in the radar plot (main Figure 4).
    Observations in main section 3.1 also apply to segmentation tasks.

\paragraph{Real Experiment Setup and Results}
In order to evaluate the sim2real transfer capability of parametric evaluation results, we curated a set of real images to perform the same evaluation. In total, we manually collected 430 images of 15 to 22 objects from various categories. For each axis, we took photo of each target object with 5 levels of the corresponding distribution shift that matches the intensity level 0, 0.25, 0.5, 0.75, 1 in the synthetic data. For example, when collecting data for a microwave object for the articulation axis, we collect 5 images of the microwave being fully closed, 25\% open, half open, 75\% open, and fully open. An example object from the real dataset (and the comparison to the most similar counterpart in simulation) is shown in . For the zoom axis, due to the limited focal length range of our real cameras, we only covered intensity levels 0 to 0.3.

\begin{figure*}[t]
\begin{center}
\includegraphics[width=\textwidth]{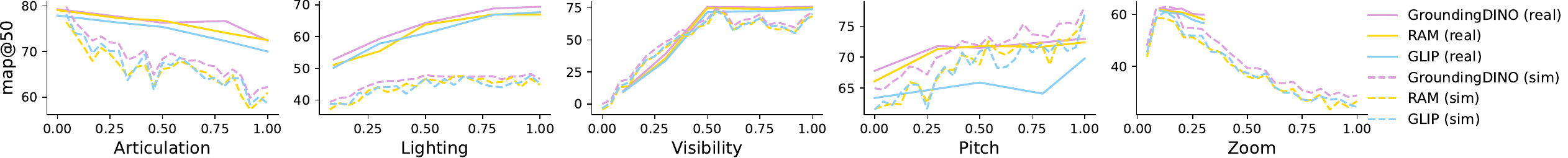}
\end{center}
    \caption{We also observe a comparable pattern in the parametric evaluation conducted on real data as observed in synthetic data (main Figure 3).}
\label{fig:real-param-eval}
\end{figure*}

We evaluated the SOTA detection methods with manually labeled bounding boxes. \Cref{fig:real-param-eval} shows that under different types of distribution shift, the performance of the SOTA methods varies on real data just as it varies in simulation.

\subsection{Holistic Scene Understanding (main paper Sec. 4.2)}
\paragraph{Details of Generation Process.}

To generate a scene traversal video, we adhere to a standard process. Initially, we sample a scene instance and subsequently define a camera trajectory using the \methodabbr toolkit. Following this, we render the traversal video, incorporating all required labels. This section will detail the specifics of the trajectory sampling procedure. In general, we want the sampled video to provide rich information (good coverage) about the scene, which can be broken down into two aspects. Firstly, we aim for the camera to physically cover the room. That means the sampled camera positions should enable visiting most open spaces in the scene, rather than just focusing on the largest open space. This guides our design for camera position sampling. Second, we want the actual video to capture as many objects in the scene as possible while still being realistic (i.e., facing the direction of movement while moving). This guides our design for camera orientation sampling. Next, we will establish the detailed steps. We will open-source all codes and generate a video dataset. 

\begin{itemize}
    \item To sample camera positions within the trajectory, a basic approach might involve randomly selecting traversable points within the scene. However, this brings the issue that points in larger rooms are more likely to be selected compared to those in smaller rooms. Our objective is to achieve a more uniform coverage across the entire scene, avoiding overconcentration in larger open areas. Thus, we used the \textbf{farthest point sampling} method to sample a set of key points that sparsely span the scene. Our focus is on ensuring the trajectory covers the main open spaces, without the necessity of navigating narrow spaces such as the gap between a cabinet and a wall. To achieve this, we perform the sampling on an \textbf{eroded} version of the traversal map. This technique effectively highlights the larger open areas in a scene while eliminating smaller gaps and corners.
    \item Now we have a set of candidate key points sampled in the scene, but a view from many of them may provide similar information about the scene (for example, two nearby points in the same room). We don't need to visit both of them in the same trajectory. To ensure efficiency and avoid redundancy, we need to select a subset of these key points while still preserving a comprehensive view of the scene (such as not excluding all points from a specific room). Our selection process begins by assessing the unique information each key point provides, specifically the objects visible from that point. We place a virtual camera at each key point and rotate it 360 degrees, recording the angle at which the maximum number of objects in the scene are visible. We note both the visible objects and their corresponding angles for each key point. 
    Next, we randomize the order of the key points and go over each point sequentially to select the points that offer additional information. If a key point reveals an object not visible from \textbf{all} previously selected points, we retain it. Otherwise, we discard it. This method results in a smaller, more efficient subset of key points — referred to as waypoints in the subsequent step — which still allows a comprehensive observation of most objects in the scene.
    \item Once we have determined the set of waypoints, our next step is to devise a trajectory that connects them in the shortest possible path. To achieve this, we frame the task as a traveling salesman problem, treating the waypoints as nodes to be visited on the scene traversal graph. In this step, to ensure the camera maintains a safe distance from walls and furniture, we slightly erode the scene traversal graph. This adjustment prevents the camera from getting too close to these obstacles, ensuring smoother navigation through the scene. 
    \item After establishing the sequence of positions in the previous step, we focus on determining the camera's orientation at each position. Our goal is to mimic the behavior of a real agent exploring the scene. Therefore, while in motion, the camera faces the direction it is moving towards. Additionally, at each waypoint, the camera 'makes a stop' and turns to an optimal angle. This angle, predetermined in an earlier step, allows the camera to capture the most objects from that viewpoint. This approach serves two purposes: firstly, to ensure that the trajectory includes keyframes or angles for optimal object visibility, and secondly, to simulate the natural behavior of an agent pausing to observe the surroundings. 
\end{itemize}

\subsection{Object States and Relations Prediction (main paper Sec. 4.3)}

\paragraph{Details of Generation Process.}  For binary object relationship prediction, we synthesized 12.5k images, each annotated with one or more of the following five labels: \texttt{open}, \texttt{close}, \texttt{ontop}, \texttt{inside}, \texttt{under}. For instance, an image might depict a toy inside an open cabinet, thus making both "inside" and "open" labels applicable. 

The image sampling process is as follows: Firstly, we select a scene and a piece of furniture to serve as the primary object in the relation (e.g., a table for placing items on top). Subsequently, we determine a plausible relationship related to this base object, with annotations provided via \methodabbr. For example, an item might be positioned \texttt{ontop} or \texttt{under} a table, but not \texttt{inside} it. Following this, we select a random object to place in the scene. This object is then integrated into the scene, employing the physical state sampling function from \methodabbr to ensure its placement aligns with the predetermined relationship. For instance, we might sample a cupcake and place it at a random location on top of the table.
    Lastly, we sample a random camera pose, ensuring the placed object is centered in the frame. We then filter out any instances where the objects of interest are not adequately visible. This procedure is repeated iteratively to compile our final dataset. 

For unary object state prediction, we generated 500 images that either consist of a \texttt{filled} or empty (not filled) container, similarly 500 images for \texttt{folded}. The sampling process for unary states is simpler -- we randomly sample a scene, then place a random container/cloth object in the scene, by 50\% chance sample a filled or folded state for the target object, then also sample a random camera pose with the target object in the center.

\paragraph{Details of Architecture Design and Hyperparameters.}

\begin{figure*}[t]
\begin{center}
\includegraphics[width=0.8\textwidth]{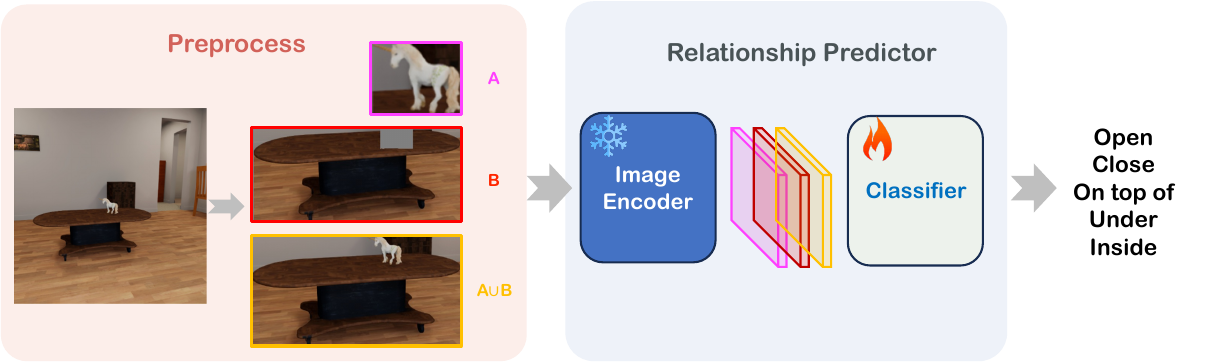}
\end{center}
    \caption{Relationship prediction model architecture used in Sec 3.3.}
\label{fig:supp-model}
\end{figure*}

Our model architecture is adapted from \cite{grounding_predicates, inayoshi2020bounding}.
Given an image input with two (or one) bounding boxes,
the model predicts the binary spatial (or unary) relationship between objects corresponding to boxes (\Cref{fig:supp-model}).
First, the model utilizes a Segment Anything image encoder \cite{kirillov2023segment} to extract hidden features.
Subsequently, RoIAlign \cite{girshick2015fast} is applied to the extracted features using the two (or one) bounding boxes.
In the binary case, where the objective is to predict the spatial relationship between the two objects, RoIAlign effectively captures spatial information from the representation

Additional features are incorporated into the aligned representations to enhance semantic information. 
Unlike \cite{grounding_predicates}, which relies on word2vec vectors \cite{mikolov2013efficient} and category names for the objects (which may not always be readily available in the world), we opt to use the Segment Anything extracted feature, from the cropped image encompassing the union of the two bounding boxes, as the additional feature. This approach preserves both spatial and semantic information.

The concatenated features are then fed into a trainable CNN to predict seven-way logits. To prevent overfitting, we freeze the Segment Anything image encoder, ensuring that the only learnable parameters are those of the randomly initialized CNN. 
Under a $0.3$ learning rate with linear scheduling, the model is trained on 13.5k synthetic images only but can achieve strong performance in the real test set (Table 4 in the main paper).

Lastly, we discuss the details of \textbf{zero-shot CLIP} \cite{radford2021learning} baseline,
which is used to mimic scenarios where synthetic datasets are not accessible. 
In this scenario, we have to rely on CLIP's zero-shot capacity.
Specifically, 
akin to our architecture, the image is cropped to maximize the semantic information.
Then the image embedding of the cropped images is compared with the label text embeddings from all verbalized prompts. 
A verbalized prompt can take the form of \texttt{<A> on top of <B>} where \texttt{<A>} and \texttt{<B>} are the placeholder for the actual object category name.
Empirically we find including the category name in the prompt outperforms using predicate only (e.g., only \texttt{on top of}).
We emphasize that having prior knowledge of category names in advance renders this task more straightforward and less fair to our approach where we have no assumption on access to any category name.

Shown in main Tables 4 and 5, 
\methodabbr is capable of generating high-quality synthetic training data by demand \cite{ge2022neural, ge2023beyond}.
Models trained on such synthetic training data are able to capture the essence of predicate prediction and bridge the sim-to-real gap.

\paragraph{Folded and Filled Prediction}
\methodabbr also supports nuanced unary object predicate such as \texttt{folded} and \texttt{filled}.
Models training on synthesized photo-realistic images can transfer well to real images.
We train two linear probes on top of the EVA02 \cite{fang2023eva} encoded representation to predict \texttt{folded} and \texttt{filled}, respectively.
We manually collect 50 real test images for each of the two predicates and observe that linear probes can achieve 86\% and 93\% real test accuracy for \texttt{folded} and \texttt{filled}, respectively.

\section{Demo Videos}
We have provided \texttt{video-demo.zip}, which consists of demo videos of our generated videos and model prediction.
Specifically, \texttt{BVS-highlight.mp4} shows all visualization content. \texttt{customizable-dataset-generator.mp4} shows scene instance augmentation. 
\texttt{predicate\_prediction.mp4} consists of predicate prediction model prediction on our collected real videos (main Sec. 4.3).
\texttt{compare-detection-\{axis\}.mp4} consists of three detection model predictions on five parametric evaluation axes (main Sec. 4.1).

\end{document}